# Large Language Models for IT Automation Tasks: Are We There Yet?

Md Mahadi Hassan[1], John Salvador[1], Akond Rahman[2], and Santu Karmaker[1]

[1]*Department of Computer Science, University of Central Florida, Orlando, FL, USA*
[2]*Department of Computer Science and Software Engineering, Auburn University, Auburn, AL, USA*

## Abstract

LLMs show promise in code generation, yet their effectiveness for IT automation tasks, particularly for tools like Ansible, remains understudied. Existing benchmarks rely primarily on synthetic tasks that fail to capture the needs of practitioners who use IT automation tools, such as Ansible. We present ITAB (IT Automation Task Benchmark), a benchmark of 126 diverse tasks (e.g., configuring servers, managing files) where each task accounts for state reconciliation—a property unique to IT automation tools. ITAB evaluates LLMs' ability to generate functional Ansible automation scripts via dynamic execution in controlled environments. We evaluate 14 open-source LLMs, none of which accomplish pass@10 at a rate beyond 12%. To explain these low scores, we analyze 1,411 execution failures across the evaluated LLMs and identify two main categories of prevalent semantic errors: failures in state reconciliation related reasoning (44.87% combined from variable (11.43%), host (11.84%), path(11.63%), and template (9.97%) issues) and deficiencies in module-specific execution knowledge (24.37% combined from Attribute & parameter (14.44%) and module (9.93%) errors). Our findings reveal key limitations in open-source LLMs' ability to track state changes and apply specialized module knowledge, indicating that reliable IT automation will require major advances in state reasoning and domain-specific execution understanding.

## 1 Introduction

DevOps practitioners rely on configuration management tools like Ansible for IT automation tasks. In order to complete these tasks, practitioners use scripts, which are referred to as automation scripts (Parnin et al., 2017). While these scripts save time and manage thousands of servers (ansible, 2022), practitioners still struggle to develop them correctly for intended IT automation tasks (Begoug et al., 2023). These challenges stem from (i) domain-specific languages with distinct syntax and semantics (Rahman et al., 2020), (ii) diverse IT automation tasks across OSes and cloud platforms (Begoug et al., 2023), and (iii) state reconciliation, which demands accurate infrastructure assessment and regulation (Hassan et al., 2024). Unsurprisingly, these challenges have sparked practitioner frustration and concern (Tanzil et al., 2023; NFsaavedra, 2024).

Given the success of large language models (LLMs) in code generation (Li et al., 2024a; Chen et al., 2021), we hypothesize that they are well-suited for automating IT tasks through automation scripts. For automation scripts, it is not enough to generate syntactically correct code; models must also accurately interpret task requirements and produce scripts that achieve the desired system state (Drosos et al., 2024) to ensure successful execution. Even minor errors such as using an incorrect module or misplacing a variable, can lead to catastrophic security risks or system failures. While benchmarks (Chen et al., 2021; Iyer et al., 2018; Odena et al., 2021) have spurred progress in code generation, they emphasize static correctness and overlook a critical question: *Can LLM-generated code actually provide solutions for IT automation tasks*? Indeed, existing approaches often lack dynamic execution testing or operate in constrained settings—for example, Ansible Wisdom (Pujar et al., 2023) relies on BLEU scores, Ansible Lightspeed (Hat, 2023) focuses on isolated tasks, and IaC-Eval (Kon et al., 2024) uses artificially generated configurations curated by human annotators. As a result, LLMs' ability to generate robust, executable Ansible scripts for real-world IT automation tasks remains largely under-explored.

To address this gap, we present **ITAB**, a benchmark for evaluating LLMs on their ability to generate *executable* IT automation scripts from real-world, user-authored natural language prompts. ITAB includes **126** tasks, carefully selected across seven key IT automation domains defined by Begoug et al. (2023): (1) *Server Configuration*, (2) *Networking*, (3) *Policy Configuration*, (4) *Templating*, (5) *Deployment Pipelines*, (6) *Variable Management*, and (7) *File Management* (details in Section 3.1). Each task must satisfy specific *operational constraints* that reflect the intended system state described by the user.

**ITAB** is different from previous work in multiple ways. First, benchmarks like IaC-Eval (Kon et al., 2024) assess whether generated code aligns with infrastructure intent specifications, while ITAB focuses on functional correctness via dynamic execution of IT automation tasks in realistic environments—marking a clear contrast with static code analysis approaches (Srivatsa et al., 2023). Second, To support reliable execution testing, we augment LLM prompts with essential context—such as file paths, configurations, and initial states—to bridge the gap between vague user instructions and executable automation scripts. Third, ITAB specifically evaluates how LLMs handle state reconciliation-a fundamental property of IT automation tools like Ansible (Hassan et al., 2024) where the orchestrator infers desired states from scripts, compares them with current states, and applies only necessary changes (Rahman and Parnin, 2023). ITAB tests this capability by validating tasks in controlled environments with predefined initial states, verifying if the resulting system state matches user requirements.

Using our test suite, we evaluated 14 open-source LLMs—selected for their accessibility, collaborative value, and cost-efficiency over proprietary models (Manchanda et al., 2024; Oketch et al., 2025)—by varying prompt specificity (TELeR levels (Santu and Feng, 2023)) and sampling temperature (Section 3.4). Overall, success rates in achieving the desired system state were strikingly low, particularly on the first attempt (pass@1), with *Templating* and *Variable Management* standing out as the most challenging IT automation domains.

Error analysis of 1,411 execution failures—instances where LLM-generated scripts did not achieve the intended system state—reveals prevalent semantic errors clustering into two fundamental categories: state reconciliation related reasoning failures (44.87%) related to how models track and manage system state across tasks, and deficiencies in module-specific execution knowledge (24.37%). We observe sampling trade-offs: higher temperatures improve diversity and pass@10 for complex scenarios, while lower temperatures enhance reliability (pass@1). Our error taxonomy provides insights for improving LLMs in IT automation and advancing research on executable reasoning. Our contributions are:

1. **Execution-driven Benchmark for IT Automation Tasks:** ITAB evaluates operational correctness, not just syntax, using real-world IT automation tasks with automated execution validation.
2. **Error Taxonomy:** We identify nine specific error categories in LLM-generated IT automation script—variable issues, host issues, path issues, attribute configuration, template errors, logic & compliance problems, module errors, output format errors, and syntax errors—establishing a standard taxonomy to reveal fundamental gaps in LLMs' ability to track state and follow instructions.

## 2 Related Works

Large Language Models (LLMs) are widely evaluated on benchmarks (Chen et al., 2021; Odena et al., 2021; Iyer et al., 2018), which focus on static code generation but overlook real-world executability. Enhanced benchmarks (Yu et al., 2024b; Zhuo et al., 2024; Jimenez et al., 2024; Yang et al., 2025; Lai et al., 2023; Zheng et al., 2025; Li et al., 2024b; Xie et al., 2024; Zhu et al., 2025; Peng et al., 2025) introduce more realistic tasks, while multilingual evaluations (Peng et al., 2024; Awal et al., 2025; Luo et al., 2025) test across languages. Semantic parsing benchmarks (Long et al., 2016; Yu et al., 2018; Yin et al., 2018a; Li et al., 2025; Yu et al., 2024a) assess natural language to code translation but often overlook system-level correctness in IT automation contexts.

Within the IT automation domain specifically, benchmarks like IaC-Eval (Kon et al., 2024) (using human-curated synthetic scenarios), WISDOM-Ansible (Pujar et al., 2023) (using BLEU scores instead of execution), and others (Khan et al., 2025; Srivatsa et al., 2024; Scheuner et al., 2014; Ragothaman and Udayakumar, 2024; Hat, 2023) evaluate LLMs on infrastructure tasks but do not include tasks that guarantees ambiguity of real-world practitioner queries. This gap is particularly problematic because IT automation requires grounding language in executable actions and system state transitions. *State reconciliation* is fundamental to tools like Ansible, where automation tools infer

```
...
  tasks:
    - name: Set fact for CentOS 7
      set_fact:
        patch_name: 'centos7-updates'
      when: ansible_distribution_major_version == '7'
    - name: Set fact for CentOS 8
      set_fact:
        patch_name: 'centos8-updates'
      when: ansible_distribution_major_version == '8'
    - name: Patch name display
      debug:
        msg: 'Patch Name {{ patch_name }}'
```

(a) Syntactically and functionally correct script.

```
...
  tasks:
    - name: Set fact for CentOS 7
      set_fact:
        patch_name: 'centos7-updates'
      when: ansible_distribution_major_version == 7
    - name: Set fact for CentOS 8
      set_fact:
        patch_name: 'centos8-updates'
      when: ansible_distribution_major_version == 8
    - name: Patch name display
      debug:
        msg: 'Patch Name {{ patch_name }}'
```
Incorrect Use, ansible_distribution_major_version is gathered as a string, not an integer.

As failed in previous two tasks, fails here too as patch_name was never set

(b) Syntactically correct but functionally incorrect script.

Figure 1: Two syntactically correct code snippets used in Ansible-based IT automation where one is functionally correct. Evaluating syntactic correctness is not enough to determine an LLM's capability for solving IT automation tasks, which requires an execution-based benchmark.

desired states, compare with current states, and apply necessary changes (Rahman and Parnin, 2023; Hassan et al., 2024). While existing studies (Wang et al., 2024a; Anandayuvaraj et al., 2024; Dou et al., 2024; Wang et al., 2024b; Chen et al., 2024) examine LLM failure patterns, they overlook challenges in reasoning about environment state and module-specific execution knowledge in IT automation.

**ITAB** addresses these gaps with executable IT automation tasks from real-world issues, enabling dynamic LLM evaluation and introducing an error taxonomy that exposes key failures by LLMs in handling state reconciliation and applying domain-specific knowledge.

## 3 Methodology

To concretely evaluate LLM-generated code against potentially ambiguous IT automation requirements, ITAB utilizes Ansible, an open-source tool where operators use YAML playbooks and task-specific modules to declaratively define a system's desired state. Assessing generation against this declarative, state-based paradigm is crucial, because syntactic correctness alone does not guarantee that a script achieves the user's intended operational outcome. Thus, ITAB frames tasks as Ansible problems where LLMs must produce playbooks that verifiably achieve a target system state, often involving state reconciliation.

### 3.1 Task Collection and Curation

To construct a benchmark with tasks that genuinely test this ability to achieve specific system states via Ansible, we began with a corpus of 52,727 Stack Overflow posts (Begoug et al., 2023) on IT Automation. Stack Overflow is a valuable resource as it contains a vast repository of user-authored questions reflecting real-world scenarios and the natural language ambiguities inherent in such problem descriptions (Yin et al., 2018b). Our benchmark development commenced with a rigorous Data Curation Phase to refine this large corpus into a high-quality set of executable candidate tasks. This phase involved two main steps:

| Topic Name | Curated Tasks |
|---|---|
| Deployment Pipelines | 15 |
| File Management | 18 |
| Networking | 16 |
| Policy Configuration | 17 |
| Server Configuration | 27 |
| Templating | 15 |
| Variable Management | 18 |
| **Total Curated** | **126** |

Table 1: Distribution of the 126 IT automation tasks across the 7 IT automation domains.

**Stratified Sampling:** To ensure our benchmark reflects the diversity of real-world Ansible usage, We first identified seven key IT automation domains based on prior analysis (Begoug et al., 2023), then applied proportionate stratified random sampling to the initial corpus to ensure topic diversity. This produced a more manageable set of "Sampled Issues" aligned with observed distributions.

**Curation & Filtering:** The "Sampled Issues" then were further refined through automated filtering to retain posts relevant to core Ansible functionalities (e.g., involving common modules like `ansible.builtin` and `community.general`) and suitable for execution. Subsequently, a rigorous manual validation stage was performed where each potential task was assessed for clarity of user intent, feasibility of implementation and validation within our defined environment.

This curation process produced **126 executable tasks** forming the IT Automation Task Benchmark (ITAB). Table 1 shows their distribution across seven IT automation domains, reflecting common Ansible use cases (Begoug et al., 2023).

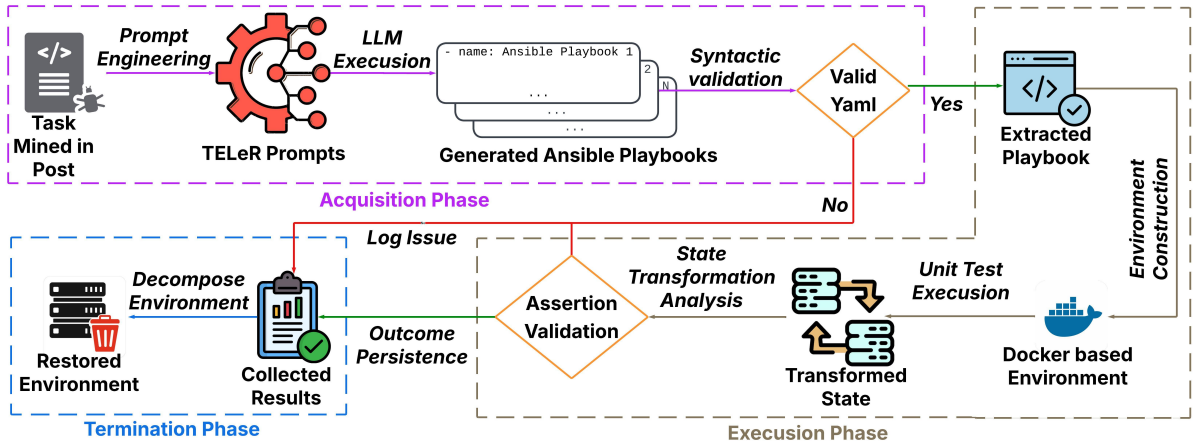

Figure 2: The ITAB evaluation pipeline, showing different phases for testing LLM-generated Ansible code.

## 3.2 Dynamic Test Case Development

We transform the curated tasks into executable test cases with validation of whether LLM-generated code achieves the intended system state. For each task in our collection, we implemented a structured transformation process:

1. **Context Analysis**: We identified the implicit system requirements, dependencies, and environmental constraints from the original question and answers. This involved careful examination of both the question text and accepted solutions to extract the underlying automation intent, often requiring domain expertise to interpret implied requirements not explicitly stated.

2. **State Definition**: We formalized both initial and target states for each task. The initial state represents the system before automation, while the target state encapsulates the desired outcome after successful execution. This step required translating often ambiguous user requirements into precise, verifiable system configurations that could be automatically validated.

3. **Parameter Identification**: We extracted key variables that might affect execution outcomes, such as file paths, service names, configuration values, and host-specific settings. This step was crucial for ensuring that our test cases could properly evaluate how LLMs handle variable substitution, path resolution, and template rendering-core capabilities for effective IT automation.

4. **Determining Functional Correctness**: For each task, we developed specific assertions to verify successful state reconciliation. These assertions checked multiple aspects of the system state including file contents, service status, and configuration values to verify the automation achieved its intended purpose. Figure 1 illustrates the distinction between syntactic and functional correctness in Ansible automation. While both scripts pass syntax validation, the right implementation fails because it incorrectly compares a string variable with integers and subsequently references an undefined variable. The left implementation correctly handles state reconciliation by using proper string comparison and default values, demonstrating why execution-based validation is essential.

To create consistent testing environments, we containerized each test scenario with precisely controlled initial states, standardized environment variables and system configurations, and created task-specific verification scripts that assess whether the desired state was achieved.

This development—led by authors with Ansible and Python expertise—produced the "Test Case Collection" for **ITAB**, comprising **733 test cases** across **126 tasks**. Our approach detects subtle failures in state reconciliation that static analysis or simple execution logging would miss. With these robust test cases in place, we next implemented a testing framework and execution pipeline to systematically evaluate LLM-generated solutions.

## 3.3 Testing Framework & Execution Pipeline

To assess operational correctness, we use a **Dynamic Validation Process** (Figure 2) that executes each generated playbook and checks it against task-specific constraints defined in Section 3.2. The pipeline consists of three main phases:

**Acquisition Phase:** This phase selects an automation issue from the test case collection and uses TELeR prompts (Section 3.4) to generate candidate playbooks via LLMs. Playbooks are then validated for YAML syntax and Ansible structure. Invalid outputs are logged and skipped; valid play-books proceed to execution.

**Execution Phase:** For each valid playbook, this phase begins by constructing a task-specific, isolated Docker environment. The playbook is then executed within this environment, triggering a transformation of system state. Our custom Python validation scripts (Section 3.2) analyze the resulting state to check whether the defined operational constraints were satisfied. Assertions based on this analysis determine the final Pass or Fail outcome.

**Termination Phase:** This phase finalizes the evaluation, whether or not the playbook was executed. Results—including Pass/Fail status, logs, and error messages—are recorded. After each task, the docker environment is reset to the default state.

### 3.4 Experimental Setup

Using the **ITAB** benchmark and evaluation pipeline (Sections 3.1–3.3), we evaluated the ability of 14 open-source LLMs (Table 12) to generate operationally correct Ansible playbooks across 126 real-world tasks. These models span sizes from 3B to 14B parameters.

We systematically varied two generation parameters: **TELeR prompt levels** and **sampling temperature**. TELeR Levels 1–3 (Santu and Feng, 2023) were used to adjust prompt specificity; higher levels encode more structured task descriptions. We focused on these levels due to the lack of few-shot examples and external documents needed for higher-level prompting. Sampling temperature (Ackley et al., 1985) was set to 0.2, 0.4, 0.6, and 0.8 to balance deterministic and exploratory behavior. For each unique (model, task, TELeR level, temperature) configuration, we generated 15 scripts to enable robust performance estimation.

Generated playbooks were evaluated using the **pass@k** metric ($k \in 1, 3, 5, 10$)(Chen et al., 2021). A sample was considered successful only if it was syntactically valid, executed correctly in our test framework (Section 3.3), and met the expected outcome. All experiments were run on uniform hardware (NVIDIA H100 GPUs).

## 4 Results and Analysis

### 4.1 Overall Performance Across Models

We evaluate LLMs' Ansible code generation using pass@k, averaged over 126 tasks, prompt styles, and temperatures (Table 2).

The results reveal the difficulty of **ITAB**: **pass@1 scores are below 4%** for nearly all mod-

| Model | @1 | @3 | @5 | @10 |
|---|---|---|---|---|
| Qwen2.5-Coder-7B-it | **3.5** | **6.7** | **8.7** | **12.0** |
| DeepSeek-Coder-V2-it | **3.1** | **5.5** | **6.7** | **8.5** |
| WizardCoder-15B | **3.1** | 3.4 | 3.6 | 3.7 |
| Llama-3.1-8B-it | 2.2 | 3.8 | 4.8 | 6.6 |
| Phi-3.5-mini-it | 2.2 | 3.4 | 4.1 | 5.0 |
| Codegemma-7B-it | 1.3 | 2.5 | 3.3 | 4.6 |
| CodeLlama-13B-it | 1.0 | 1.6 | 2.1 | 3.0 |
| Starcoder2-7B | 0.8 | 1.8 | 2.6 | 3.9 |
| Vicuna-7B-it | 0.5 | 1.1 | 1.5 | 2.3 |
| Qwen2.5-7B-it | 2.0 | 3.3 | 4.0 | 5.0 |
| CodeLlama-7b-it | 0.9 | 1.3 | 1.6 | 2.2 |
| Llama-3.2-3B-it | 1.1 | 1.6 | 2.0 | 2.7 |
| DeepSeek-Distill-L | 0.7 | 0.9 | 1.0 | 1.2 |
| DeepSeek-Distill-Q | 0.2 | 0.6 | 0.7 | 0.8 |

Table 2: pass@k for selected LLMs on **ITAB** (avg. over tasks, prompts, temps).

els, showing that reliably generating correct automation script on the first attempt remains a major challenge—even with environment context. While pass@10 improves slightly (Qwen-Coder at 12.0%), overall success remains modest, highlighting the gap between syntactic fluency and true operational correctness in IT automation tasks.

Pretraining data composition affects performance, with top models like Qwen2.5-Coder-7B (70% code/20% text (Hui et al., 2024; Yang et al., 2024)) and DeepSeek-Coder-V2 (60% code/30% text (Zhu et al., 2024)) effectively combining code and language input—key for understanding prompts and generating correct Ansible. Qwen2.5-Coder-7B's edge over its base model underscores how code specialization benefits from strong natural language grounding.

The general-purpose Llama-3.1-8B (15T+ tokens (Dubey et al., 2024)) outperformed code-specialized CodeLlama models (e.g., 13B, 85% code/500B tokens (Rozière et al., 2023)), suggesting massive scale and language understanding can be more vital than high code ratios for instruction-grounded IT automation tasks. These trends hint that success in ITAB might be influenced not only by code volume but also by the interplay of code exposure, general language understanding, and dataset scale.

### 4.2 Impact of Temperature and Prompting

Beyond overall success rates, this section analyzes how generation configurations—specifically sampling temperature and prompt detail—influence LLM performance on IT automation tasks.

**Sampling Temperature:** Sampling temperature significantly impacted performance, revealing a clear trade-off. As shown in Figure 3 (a), lower temperatures (e.g., 0.2) maximized pass@1

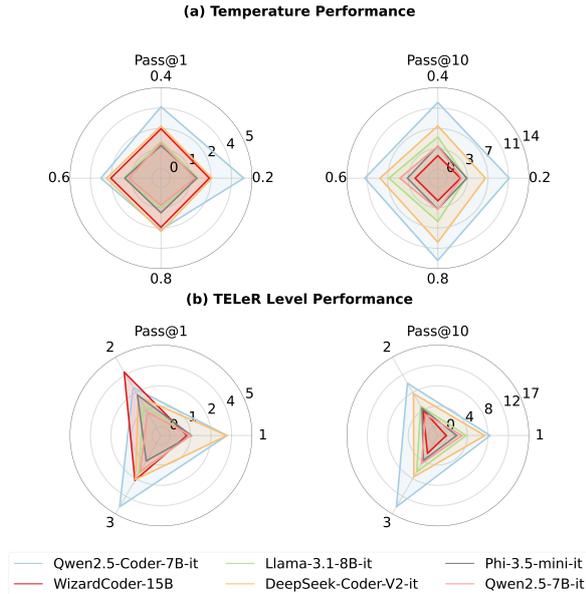

Figure 3: Impact of decoding parameters for the top 6 LLMs. (a) Pass@1 and Pass@10 across temperature settings. (b) Pass@1 and Pass@10 across TELeR levels. A greater distance from the center indicates higher success rates and all values are shown as percentages.

scores by favoring reliable, deterministic outputs. Conversely, higher temperatures (0.6–0.8) boosted pass@10 by increasing output diversity, aiding discovery in complex tasks. This highlights a practical precision-versus-exploration dilemma for tuning generation.

**Prompt Detail (TELeR Levels):** While Figure 3 (b) shows higher TELeR levels often boost Pass@10 accuracy for capable models like Qwen2.5-Coder-7B by leveraging richer input, increased prompt detail is not uniformly advantageous. It can also result in overly complex or 'over-engineered' solutions (Section 6), with the overall benefit varying by model capability.

| Model | @1 | @3 | @5 | @10 |
|---|---|---|---|---|
| Qwen2.5-Coder-7B-it | **5.6** | **9.8** | **12.1** | **15.5** |
| Phi-3.5-mini-it | **4.2** | **6.3** | **7.3** | **8.3** |
| DeepSeek-Coder-V2-it | 3.3 | 5.5 | 6.6 | 8.3 |
| Llama-3.1-8B-it | 2.8 | 4.8 | 5.9 | 7.5 |
| Codegemma-7B-it | 2.6 | 4.4 | 5.4 | 7.0 |
| Starcoder2-7B | 0.8 | 1.9 | 2.6 | 3.8 |
| Vicuna-7B-it | 0.5 | 1.1 | 1.4 | 2.0 |
| WizardCoder-15B | 2.7 | 2.9 | 3.2 | 3.4 |
| CodeLlama-13B-it | 2.7 | 4.9 | 6.1 | 8.1 |

Table 3: pass@k with error-avoidance prompts (avg. across configs).

**Error-Aware Prompting:** To test if targeted guidance helps, we used error-aware prompts informed by our failure taxonomy (Section 5), including hints against common mistakes. However, Table 3 shows this yielded only marginal pass@k improvements (1–4 percentage points), indicating that prompt modifications alone poorly mitigate models' core challenges in state reconciliation reasoning and module-specific knowledge.

### 4.3 Performance by IT Automation Domains

Figure 4 analyzes LLM performance across IT automation domains, with rows representing domains, columns as model families, and color intensity indicating pass@1 or pass@10 rates.

While models performed better on routine tasks like Server Configuration and File Management, categories requiring precise state and variable handling—notably Templating and Variable Management—proved far more challenging, exhibiting the lowest pass@k scores. This suggests these state-sensitive domains demand reasoning beyond basic syntax or module use, aligning with our error analysis (Section 5) where difficulties in state reconciliation related issues are prevalent.

## 5 Error Taxonomy

As we saw the low pass@k values, we wanted to investigate what really lies behind the numbers. We conducted an extensive qualitative study of 1,411 execution failures spanning across IT automation domains and models, creating a taxonomy of errors in open-source models generating Ansible script for real-world issues.

Our error analysis reveals striking patterns. Notably, models distilled for enhanced reasoning overwhelmingly fail at basic syntax—suggesting that general reasoning does not readily transfer to structured code domains like IT automation without explicit domain grounding. For the other evaluated LLMs, failures predominantly extend beyond syntax and cluster into two fundamental semantic categories: First, deficiencies in module-specific execution knowledge are prevalent. These models may identify appropriate Ansible modules but frequently err in their precise implementation, with Attribute and Parameter Errors being common. This highlights a gap where models understand what automation action is needed but not how to correctly configure module specifics. Second, failures in state reconciliation related reasoning are widespread. This encompasses errors in variable handling and path navigation—underscoring difficulties in tracking state across complex scopes (playbooks, roles, templates)—and flawed template (Jinja2) logic, indicating problems reasoning about dynamic content generation. The distribution of these state-related errors (including variable, host, path, and template issues) also varies by model architecture, suggesting differing blindspots in contextual reasoning.

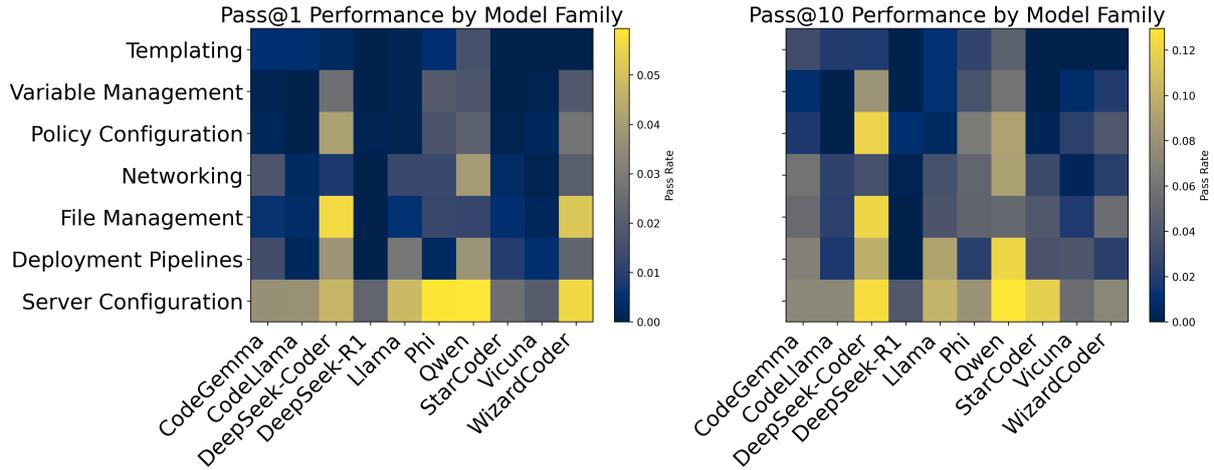

Figure 4: Pass@1 and Pass@10 Performance Across IT Automation Domain Clusters for Four Model Families

| Model | Variable issue | Host Issues | Path Issues | Attribute & parameter Errors | Template Issues | Logic & Compliance | Module Errors | Output Format | Syntax |
|---|---|---|---|---|---|---|---|---|---|
| Codegemma-7B-it | **21.951** | 3.659 | **24.39** | 18.293 | 10.976 | 4.878 | 4.878 | 2.439 | 8.537 |
| CodeLlama-7b | **19.178** | 8.219 | 15.068 | 6.849 | **23.288** | 5.479 | 8.219 | 4.11 | 8.219 |
| CodeLlama-13B | 8.654 | 9.615 | 10.577 | 10.577 | 4.808 | 3.846 | 5.769 | 0.962 | **45.192** |
| DeepSeek-Coder-V2 | 1.695 | **32.203** | 20.339 | 10.169 | 10.169 | 3.39 | 13.559 | 0 | 8.475 |
| Llama-3.1-8B | 3.333 | 13.333 | 11.111 | 11.111 | 4.444 | 7.778 | 17.778 | 5.556 | **26.667** |
| Llama-3.2-3B | 13.043 | 16.304 | 8.696 | 14.13 | 3.261 | 3.261 | 15.217 | 0 | **26.087** |
| Phi-3.5-mini | **26.374** | 8.791 | 15.385 | **16.484** | 14.286 | 6.593 | 10.989 | 0 | 2.198 |
| Qwen2.5-7B | 7.216 | 8.247 | 10.309 | **22.68** | 5.155 | 7.216 | **19.588** | 2.062 | 17.526 |
| Qwen2.5-Coder-7B | 13.514 | 7.207 | 11.712 | **19.82** | 8.108 | 0.901 | 5.405 | **17.117** | 16.216 |
| Starcoder2-7B | 4.938 | 0 | 6.173 | 25.926 | 11.111 | 11.111 | 13.58 | 1.235 | **28.395** |
| Vicuna-7B | 10.417 | 8.333 | 0 | 10.417 | 10.417 | **14.583** | 4.167 | 0 | **41.667** |
| WizardCoder-15B | 6.796 | **26.214** | 5.825 | 6.796 | 13.592 | 0.971 | 0 | 3.883 | **28.155** |
| DeepSeek-Distill-L | 0 | 1.19 | 1.19 | 5.952 | 1.19 | 3.571 | 4.762 | 0 | **79.762** |
| DeepSeek-Distill-Q | 0 | 0 | 0 | 0 | 0 | 0 | 0 | 0 | **100** |
| Aggregation | 11.43 | 11.84 | 11.63 | 14.44 | 9.97 | 5.83 | 9.93 | 3.114 | 21.45 |

Table 4: Distribution of Error Types (%) Across Models. Bold values indicate the most frequent error category per model. To avoid skew, syntax-heavy error distributions from reasoning-distilled models are excluded from the aggregated percentages.

This analysis reveals key gaps—from basic syntax issues in some models to deeper failures in state reasoning and module-specific knowledge—explaining the low pass@k performance on executable IT automation tasks.

## 6 Qualitative Examples

To complement our quantitative results, we qualitatively analyze Qwen-Coder—the top-performing model—on representative automation tasks. For instance, examining the impact of decoding parameters revealed interesting trade-offs.

In the targeted host execution (centos1) task, increasing sampling temperature (T) broadened output diversity and the variety of errors, with distinct error categories rising from four at T=0.2 to eight at T=0.8 (Table 5). Temperature also played a key role in discovering specific correct behavior: in the check mode execusion task, only samples at T=0.8 correctly applied check_mode in two instances, example in Listing 1, which was absent at lower temperatures. On the *multi-server directory setup* task, more detailed prompts (TELeR Level 3) produced correct but overengineered solutions, including redundant file creation and validations that could reduce maintainability.

| Error Type | T=0.2 | T=0.8 |
|---|---|---|
| Attribute type mismatch | 15 | 15 |
| Invalid YAML | 13 | 9 |
| Wrong host name | 3 | 7 |
| Conflicting action | 5 | 1 |
| Variable issue | 0 | 1 |
| Wrong attribute | 0 | 1 |
| Template error | 0 | 1 |
| Missing hosts field | 0 | 2 |

Table 5: Error Diversity by Temperature for 'Targeted host execution'.

These case studies show that even top models are sensitive to temperature, prompt design, and task complexity—higher temperatures boost exploration but increase errors, while detailed prompts aid correctness but can add unnecessary complexity.

```
...
tasks: ...
- name: Send email if any host changes
  mail:
     host: localhost
     ...
  when: mail_result.rc == 0 and
     ansible_check_mode
```

Listing 1: Use of Check Mode to Control Email Sending

## 7 Discussion

Evaluation using **ITAB** reveals significant challenges for current open-source LLMs in generating operationally correct Ansible code, highlighting a critical gap between syntactic validity and reliable execution in complex, stateful IT automation.

Extremely low success rates (pass@10 max 12.0%) reveal LLMs struggle with functional correctness for real-world instructions, even with context. Our analysis indicates that models pretrained on extensive code and substantial natural language outperform those with less balanced data. This suggests success in IT automation demands not just coding ability but also robust interpretation of complex user instructions.

Furthermore, our analysis of generation parameters showed that while sampling temperature provides a crucial lever for balancing reliability (**pass@1**) and exploration (**pass@10**). prompt engineering yielded limited net improvements. While higher TELeR levels offered some multi-sample gains for certain models, this was often offset by increased solution complexity. Similarly, incorporating error-aware guidance improved performance *significantly*. This suggests that even with stronger prompts, models struggle due to more fundamental limitations—such as reasoning about system state, interpreting variable scopes, and adhering to procedural constraints.

These difficulties-especially in managing variable resolution and template logic-manifest clearly in task-specific performance and error patterns. Models consistently struggled with IT automation domains requiring precise state and variable handling, namely **Templating** (using Jinja2) and **Variable Management** (Section 4.3). Our error analysis confirms this pattern, revealing that conventional LLMs primarily fail due to state reconciliation reasoning issues (44.7% of errors across variable, host, path, and template categories) and limited module-specific execution knowledge (24.37%), confirming that tracking state changes and applying domain-specific knowledge are primary bottlenecks.

Our findings imply that current open-source LLMs exhibit critical deficits in state reasoning and precise execution, hindering reliable IT automation. Overcoming key bottlenecks, such as variable and template management, demands more than prompt tuning—pointing to needs for architectural or domain-specific enhancements. Consequently, the observed high failure rates mandate dynamic execution validation, as ITAB provides, since static checks alone cannot ensure operational safety and correctness.

On the brighter side, **ITAB** introduces a novel and challenging benchmark—difficult even for leading open-source models like DeepSeek and Qwen—laying the foundation for future research in AI-powered IT automation, including advances in model architectures, fine-tuning strategies, and domain-specific reasoning.

## 8 Conclusion

Ensuring the reliability of Large Language Models for complex IT automation tasks remains a significant challenge, largely due to the limitations of existing evaluation methods. We presented **ITAB**, a benchmark leveraging real-world practitioner queries and dynamic execution, to rigorously assess the functional correctness of LLM-generated Ansible scripts. Our comprehensive evaluation of 14 open-source LLMs revealed strikingly low success rates in achieving desired system states, with predominant failures stemming not merely from syntax but from critical deficiencies in state reconciliation reasoning and the application of module-specific execution knowledge. These fundamental bottlenecks, which prompt engineering alone proved insufficient to resolve, underscore that current open-source LLMs are not yet consistently reliable for autonomous IT automation. Significant advances in models' stateful reasoning and domain adaptation capabilities are therefore necessary. We posit that challenging, execution-based benchmarks like ITAB are essential for guiding and measuring future progress towards truly dependable language-guided automation that correctly achieves the intended system state.

## 9 Limitations

While ITAB provides valuable insights into LLMs' capabilities for IT automation tasks, several limitations should be acknowledged. First, our evaluation focuses exclusively on open-source models ranging from 3-14B parameters, leaving questions about how larger proprietary models or smaller specialized models might perform. Second, our benchmark's sample size (126 tasks) represents a limited subset of the vast IT automation landscape. While carefully curated to cover diverse categories, this sample may not capture all edge cases or specialized scenarios. Finally, our evaluation primarily tests individual playbooks rather than complex multi-playbook orchestration scenarios that are common in enterprise environments.

## A Task Curation Details

This appendix provides supplementary details on the curation process for the **ITAB** benchmark tasks described in Section 3.1.

### A.1 Initial Filtering Criteria

Posts from the initial Stack Overflow corpus (Begoug et al., 2023) were automatically filtered based on:

1. **Replicability:** The described problem needed to be potentially replicable within a containerized Linux environment, excluding issues specific to non-containerizable hardware, proprietary systems unavailable in Docker, or GUI interactions.
2. **Core Ansible Modules:** The problem or its likely solution needed to involve modules primarily from the `ansible.builtin`, `ansible.netcommon`, `ansible.utils`, or `community.general` collections, focusing on common, well-supported functionalities. Posts relying heavily on obscure, deprecated, or highly specialized external collections were typically excluded.

### A.2 IaC Topic Descriptions

The seven IaC topics used for categorization, based on (Begoug et al., 2023), encompass:

1. **Server Configuration:** Managing server environments, package installation/removal, service states (start, stop, restart), user/group management, basic system settings.
2. **Policy Configuration:** Defining security settings (firewalls, SELinux), compliance rules, access controls, system policy enforcement.
3. **Networking:** Configuring network interfaces, IP addressing, routing, DNS, network services (DHCP, NTP), basic network device interactions.
4. **Deployment Pipelines:** Automating steps in application deployment, including fetching artifacts, managing dependencies, deploying code, database migrations, basic CI/CD tasks.
5. **Variable Management:** Handling variable definition, scope (host, group, play), precedence, lookup plugins, complex data structures, and accessing facts.
6. **Templating:** Using dynamic templates (primarily Jinja2) to generate configuration files based on variables and logic.
7. **File Management:** Creating, modifying (lineinfile, blockinfile, replace), deleting files and directories, managing permissions, copying/fetching files.

### A.3 Manual Curation Criteria

The set of posts selected via stratified sampling (initially 200 posts) underwent manual review. Posts were excluded during this final curation if:

1. **Feasibility Issues:** Upon closer inspection, the task required external services, hardware,

credentials, or network configurations impractical or insecure to replicate reliably within the isolated Docker test environment.

2. **Ambiguity/Incompleteness:** The problem description was too vague, lacked crucial details, or the accepted SO solution was unclear, incomplete, or non-functional, preventing the formulation of a clear, testable task and validation criteria.

3. **Relevance Issues:** The post, despite keywords, did not ultimately represent a core Ansible automation task solvable via the targeted modules or involved primarily debugging Ansible itself rather than using it for automation.

This rigorous process yielded the final 126 tasks for benchmark development.

## B Docker-based Execution Environment

To ensure consistent, isolated, and reproducible execution for both reference solutions and LLM-generated playbooks, we designed a standardized Docker-based testing environment. This environment simulates a small, heterogeneous infrastructure network. We configured a dedicated Docker network with the subnet `10.1.1.0/24` and assigned the gateway address `10.1.1.254`. Within this network, we provisioned four distinct Docker containers acting as target nodes, each running a common Linux distribution frequently encountered in production systems:

- `ubuntu1`: Ubuntu Linux (IP: `10.1.1.1`)
- `alpine1`: Alpine Linux (IP: `10.1.1.2`)
- `centos1`: CentOS Linux (IP: `10.1.1.3`)
- `redhat1`: Red Hat Enterprise Linux (or a compatible distribution like Rocky/Alma Linux) (IP: `10.1.1.4`)

Each container was equipped with SSH access and Python, prerequisites for Ansible control. The specific target node(s) for each benchmark task were determined by the task's requirements, often specified implicitly (e.g., tasks involving 'apt' target Ubuntu, 'yum'/'dnf' target CentOS/Red Hat) or explicitly in the adapted task description. This multi-node, multi-distribution setup allows for evaluating the correctness and portability of Ansible playbooks across diverse target systems, ensuring generated solutions are robust and not overly fitted to a single environment. Each test execution (detailed in Section 3.3) occurs within a fresh instance of this environment.

## C Decoding Strategies

We systematically varied two key aspects influencing LLM generation: the structure of the input prompt and the sampling temperature.

### C.1 Prompt Design based on TELeR Level of Detail

To explore how the amount of information provided in the prompt affects Ansible code generation, we adapted the TELeR taxonomy (Santu and Feng, 2023), which categorizes prompts along dimensions including Turn, Expression, Level of Detail, and Role. Our study focused specifically on the **Level of Detail** dimension. This dimension describes the richness of the input prompt, ranging from minimal instruction (Level 0) to incorporating external documents (Level 5) or requesting explanations (Level 6).

Based on our objective of evaluating core Ansible generation capabilities and the nature of our benchmark tasks (derived from SO posts without readily available ideal solutions for few-shot examples or relevant external documents for RAG), we constrained our prompt variations to specific TELeR levels:

- **Excluded Levels:** We excluded Level 0 (minimal detail, likely insufficient guidance), Level 4 (requiring few-shot examples or evaluation guidelines not suitable for our setup), Level 5 (requiring external documents for RAG, which were not available per task), and Level 6 (focused on explainability, not core generation).
- **Utilized Levels:** Consequently, our experiments employed three prompt variations corresponding to **TELeR Levels** 1, 2, and 3. These levels represent increasing detail within the prompt itself, ranging from basic instruction (Level 1) to more elaborated descriptions potentially including constraints or hints derived from the stack overflow post analysis (Level 3), without relying on external examples or documents. Those prompts are presented in Table 6-8.
- **Error aware prompts:** We also experimented with how much error-aware prompts affect the LLM generation. We inject the errors to avoid that we got from Error Taxonomy (Section 5). We inject the error information in all three level of prompts presented in Table 9- 11.

This systematic variation allows us to assess the sensitivity of different LLMs to the level of detail provided directly within the prompt for Ansible task generation.

## C.2 Temperature Variation

The temperature parameter significantly influences the randomness and creativity of LLM outputs during sampling (Ackley et al., 1985; OpenAI, 2025). Lower temperatures (closer to 0) make the model's output more deterministic, favoring the highest probability tokens at each step, leading to more focused but potentially less diverse results. Higher temperatures (e.g., approaching 1.0) increase randomness, allowing the model to sample lower probability tokens more often, potentially leading to more novel or diverse solutions but also increasing the risk of errors or nonsensical output (Shivam Mehta, 2023).

To understand the trade-off between correctness and diversity for Ansible code generation, we systematically generated samples using four distinct temperature settings: **0.2, 0.4, 0.6, and 0.8**. This range allows us to observe performance from near-deterministic generation (0.2) to significantly more stochastic generation (0.8). The impact of temperature is analyzed in Section 4.

## C.3 Experiment Design

### C.3.1 Evaluation Metrics

We evaluate the functional correctness of generated Ansible playbooks using the **pass@k** metric (Chen et al., 2021). This metric estimates the probability that at least one of the top $k$ generated samples for a task passes all validation criteria. A sample is considered "correct" ($c$) only if it is syntactically valid, executes successfully without errors, achieves the desired functional outcome verified by our task-specific validation tests (Section 3.2), and is idempotent, all within the testing framework described in Section 3.3.

We use the unbiased estimator for pass@k, averaged across all benchmark tasks:

$$pass@k = \mathbb{E}_{\text{Tasks}}\left[1 - \frac{\binom{n-c}{k}}{\binom{n}{k}}\right] \quad (1)$$

where $n$ is the total number of samples generated per task and $c$ is the count of correct samples. In our experiments, we report results for $k \in \{1, 3, 5, 10\}$. We generated $n = \langle\text{Specify your value of } n\rangle$ samples per task for each LLM configuration (Section 3.4) to ensure robust estimation. While pass@k is the primary metric, analysis may also consider intermediate failure points like syntax errors.

### C.3.2 LLM Selection and Configuration

The list of LLMs and their versions are given in Table 12. We evaluated a diverse set of fourteen large language models (LLMs), including both code-specialized models (Code LLMs) and general-purpose instruction-tuned models (General LLMs). Our selection spans multiple model families, parameter scales, and release dates to capture variations in architecture, training data, and objectives. Table 12 lists the specific models used in our study along with their key characteristics.

Each model was prompted to generate solutions for the 126 Ansible benchmark tasks (Section 3.2). For every task, we explored variations in the generation process using multiple prompting strategies and sampling temperatures, **as detailed in Section 3.4**. To compute the *pass@k* metrics (Section C.3.1), we generated $n = 15$ code samples for each unique combination of task, model, prompt style, and temperature setting.

### C.3.3 Experimental Execution

The inference process for generating Ansible code samples from all selected Large Language Models (LLMs) was conducted on a single NVIDIA H100 GPU to ensure hardware consistency. For each of the 14 LLMs (Section 3.4), we generated responses for every one of the 126 benchmark tasks (Section 3.2).

As detailed in Section 3.4, we tested three distinct prompt structures (derived from TELeR Levels 1-3) and four different sampling temperature settings (0.2, 0.4, 0.6, 0.8). For each unique combination of model, task, prompt style, and temperature setting, we generated $n = 15$ independent samples, as required for the *pass@k* evaluation (Section C.3.1).

This resulted in a substantial number (22,680) of generated IT automation script per model. Where On average Each model needed 20 GPU hours to generate all those scripts. We user NVIDIA H100 GPU having 80GB memory to generate scripts from LLMs. So go generate the whole results we needed 280 GPU hours (Approximately).

Executing all experiments on uniform hardware ensures that observed performance differences can be attributed to the models and prompting strategies themselves, enabling a fair comparison across all configurations. The execution and evaluation of each generated sample followed the pipeline described in Section 3.3.

| Component | Content |
| --- | --- |
| **system_role** | As an expert Ansible developer and Linux systems administrator, your role is to analyze Stack Overflow posts and transform them into practical, self-contained, and well-documented solutions while ensuring cross-distribution compatibility, adhering to infrastructure-as-code best practices, considering distribution-specific differences, validating network requirements, ensuring proper YAML syntax, and verifying completeness without assuming external dependencies or pre-existing configurations. |
| **prompt** | Transform the given Stack Overflow post, including its title ({{title}}) and body ({{body}}), into a production-ready Ansible playbook solution that adheres to the specified target environment, which consists of a network with a subnet of 10.1.1.0/24 and a gateway at 10.1.1.254, along with four nodes—ubuntu1 (10.1.1.1) running Ubuntu Linux, alpine1 (10.1.1.2) running Alpine Linux, centos1 (10.1.1.3) running CentOS Linux, and redhat1 (10.1.1.4) running Red Hat Linux—while ensuring cross-distribution compatibility, complying with the implementation constraint ({{constraint}}), and delivering a self-contained, production-ready playbook that includes a brief analysis of the problem and solution approach, incorporates all necessary variables and files, follows YAML best practices, features comprehensive error handling, and is enclosed in triple backticks ("'") for automated Python processing. |

Table 6: Level 1 TELeR (Santu and Feng, 2023) Prompt Structure for IT Automation Tasks. Placeholder expressions enclosed in double curly braces (e.g., {{title}}, {{body}}, {{constraint}}) indicate variable components that will be dynamically replaced with task-specific information.

### C.3.4 NLP packages

We use the following python packages for building our experiments: openai, tiktoken, accelerate, ansible, ansiblemetrics, colorama, datasets, ipython, numpy, pandas, tokenizers, torch, tqdm, transformers, argcomplete, simple_term_menu, paramiko, scp, pynvml, scipy, bitsandbytes, openpyxl, docker, bs4, protobuf, sentencepiece, pymysql, natsort, umap-learn, sentence-transformers, hdbscan.

## D Qwen 2.5 Coder 7B Model Case Study

Our detailed case study of Qwen2.5-Coder-7B, the top-performing model in our evaluation, revealed several nuanced behaviors and challenges. Initial observations indicated inconsistent performance trends related to sampling temperature, with some tasks showing improved success at higher temperatures (an "upward trend") while others performed better at lower temperatures (a "downward trend"). Similarly, the impact of TELeR prompt levels was not uniform across all tasks, prompting further investigation into when increased prompt detail was genuinely beneficial versus when it was not effectively utilized. The 'Templating' domain was identified as consistently underperforming for this model, suggesting that the task constraints or the model's approach to this category might require specific attention.

Further investigation into these patterns provided deeper insights. A key observation was that on higher TELeR levels of prompt detail, Qwen2.5-Coder-7B tended to generate overly complicated or 'over-engineered' playbooks. In some instances, these more complex Level 3 responses, while potentially containing elements of a correct solution, ultimately failed due to issues introduced by this over-complication. We also observed instances where the model appeared to treat code snippets or patterns directly from Stack Overflow posts as factual or directly applicable without sufficient adaptation to the specific task constraints presented in our benchmark. These findings suggest that even for high-performing models, achieving reliable and minimally complex solutions may require very clear, unambiguous instructions, potentially needing to 'spoon-feed' critical details to avoid misinterpretation or over-complication.

To further understand these temperature-related performance trends, we conducted an error message analysis. This study revealed several key observations regarding how sampling temperature influences solution diversity and success patterns. We found that any upward trend in performance with increasing temperature predominantly occurred

| Component | Content |
| --- | --- |
| **system_role** | As an expert Ansible developer and Linux systems administrator, you possess deep knowledge of various Linux distributions, networking, and infrastructure automation. Your role involves analyzing Stack Overflow posts and transforming them into practical Ansible solutions while ensuring cross-distribution compatibility in playbooks and adhering to infrastructure-as-code best practices. When approaching problems, you should think step-by-step, carefully considering distribution-specific differences, validating network requirements, ensuring proper YAML syntax, and verifying the completeness of each solution. Assumptions about external dependencies or pre-existing configurations should be avoided, as all solutions must be self-contained and well-documented. |
| **prompt** | Transform the given Stack Overflow post, including its title (title) and body (body), into a production-ready Ansible playbook solution that adheres to the specified target environment and implementation constraint. The target environment consists of a network with a subnet of 10.1.1.0/24 and a gateway at 10.1.1.254, along with four nodes: ubuntu1 (10.1.1.1) running Ubuntu Linux, alpine1 (10.1.1.2) running Alpine Linux, centos1 (10.1.1.3) running CentOS Linux, and redhat1 (10.1.1.4) running Red Hat Linux. The implementation must comply with constraint: 'constraint' while ensuring cross-distribution compatibility. The deliverable should include a brief analysis of the problem and the solution approach, along with a fully self-contained, production-ready Ansible playbook that functions across all specified distributions. The playbook must incorporate all necessary variables and files, feature comprehensive error handling, and adhere to YAML best practices. Additionally, it must be enclosed in triple backticks ("') for automated Python processing. The primary focus is on creating a robust, directly implementable solution for the target environment. |

Table 7: Level 2 TELeR (Santu and Feng, 2023) Prompt Structure for IT Automation Tasks. Placeholder expressions enclosed in double curly braces (e.g., {{title}}, {{body}}, {{constraint}}) indicate variable components that will be dynamically replaced with task-specific information.

| Component | Content |
|---|---|
| **system_role** | You are an expert Ansible developer and Linux systems administrator with deep knowledge of different Linux distributions, networking, and infrastructure automation. Your role is to:<br>1. Analyze Stack Overflow posts and transform them into practical Ansible solutions<br>2. Ensure cross-distribution compatibility in playbooks<br>3. Follow infrastructure-as-code best practices<br>You should:<br>- Think step-by-step when analyzing problems<br>- Consider distribution-specific differences<br>- Validate network requirements<br>- Ensure proper YAML syntax<br>- Verify solution completeness<br>Never assume external dependencies or pre-existing configurations. All solutions should be self-contained and well-documented. |
| **prompt** | Transform the following Stack Overflow post into an Ansible playbook solution.<br>STACK OVERFLOW POST:<br>Title: title<br>Body: body<br>TARGET ENVIRONMENT:<br>Network:<br>- Subnet: 10.1.1.0/24<br>- Gateway: 10.1.1.254<br>Nodes:<br>1. ubuntu1 (10.1.1.1) - Ubuntu Linux<br>2. alpine1 (10.1.1.2) - Alpine Linux<br>3. centos1 (10.1.1.3) - CentOS Linux<br>4. redhat1 (10.1.1.4) - Red Hat Linux<br>IMPLEMENTATION CONSTRAINT: constraint<br>DELIVERABLE REQUIREMENTS:<br>1. Brief analysis of the problem and your solution approach<br>2. Production-ready, self-contained Ansible playbook that:<br><br>• Works across all specified distributions<br><br>• Includes all necessary variables and files<br><br>• Features comprehensive error handling<br><br>• Follows YAML best practices<br><br>3. Must be enclosed in triple backticks ("'") for automated Python processing<br>Note: Focus on creating a robust, production-ready playbook that can be directly implemented in the target environment. |

Table 8: Level 3 TELeR (Santu and Feng, 2023) Prompt Structure for IT Automation Tasks. Placeholder expressions enclosed in double curly braces (e.g., {{title}}, {{body}}, {{constraint}}) indicate variable components that will be dynamically replaced with task-specific information.

| Component | Content |
|---|---|
| **system_role** | You are an expert Ansible developer and Linux administrator; always generate playbooks that comply with given constraint, avoid undefined variables and invalid paths, and carefully select modules appropriate for the specified constraint and runtime environment and use only its appropriate attributes, ensuring valid YAML and correct Ansible syntax, and follow the output pattern in constraint. |
| **prompt** | Transform the given Stack Overflow post, including its title (title) and body (body), into a production-ready Ansible playbook solution that adheres to the specified target environment, which consists of a network(subnet- 10.1.1.0/24) with four nodes—ubuntu1, alpine1, centos1, and redhat1, complying with the implementation constraint (constraint), and delivering a self-contained, production-ready playbook, incorporates all necessary variables and uses correct paths, follows YAML best practices, and is enclosed in triple backticks for automated Python processing. |

Table 9: Level 1 TELeR (Santu and Feng, 2023) Prompt Structure with error awareness for IT Automation Tasks. Placeholder expressions enclosed in double curly braces (e.g., {{title}}, {{body}}, {{constraint}}) indicate variable components that will be dynamically replaced with task-specific information.

| Component | Content |
|---|---|
| **system_role** | As a senior Ansible automation engineer with deep Linux expertise, you generate robust, production-ready playbooks for diverse distributions. Always follow best practices, strictly comply with constraint. Make sure to avoid using undefined variables and invalid file paths. Select modules that are suitable for the given constraints and runtime environment, and use only their correct attributes. Ensure that the generated YAML is valid and free from Ansible syntax or logic errors. The output follows the specified format outlined in the constraint. |
| **prompt** | Transform the given Stack Overflow post, including its title (title) and body (body), into a production-ready Ansible playbook solution that adheres to the specified target environment and implementation constraint. The target environment consists of a network with a subnet of 10.1.1.0/24 and a gateway at 10.1.1.254, along with four nodes: ubuntu1 (10.1.1.1), alpine1 (10.1.1.2), centos1 (10.1.1.3), and redhat1 (10.1.1.4). The implementation must comply with constraint: 'constraint'. The deliverable should include a brief analysis of the problem and the solution approach, along with a fully self-contained, production-ready Ansible playbook that functions across all specified distributions. Make sure to avoid using undefined variables and invalid file paths. Select modules that are suitable for the given constraints and runtime environment, and use only their correct attributes. Ensure that the generated YAML is valid and free from Ansible syntax or logic errors. The output follows the specified format outlined in the constraint, and adhere to YAML best practices. Additionally, it must be enclosed in triple backticks ("`") for automated Python processing. The primary focus is on creating a robust, directly implementable solution for the target environment. |

Table 10: Level 2 TELeR (Santu and Feng, 2023) Prompt Structure with error awareness for IT Automation Tasks. Placeholder expressions enclosed in double curly braces (e.g., {{title}}, {{body}}, {{constraint}}) indicate variable components that will be dynamically replaced with task-specific information.

| Component | Content |
| --- | --- |
| **system_role** | You are an expert Ansible developer. When generating playbooks, always:<br>1. Strictly comply with given constraint (including output format, filenames, and requirements)<br>2. Define all variables before use and avoid undefined variable errors<br>3. Avoid using invalid file paths.<br>4. Carefully examine the constraint and runtime environment, and select modules appropriate for both.<br>5. Use only valid attributes for the selected modules.<br>6. Ensure valid YAML and correct Ansible syntax and structure.<br>7. Avoid logic errors.<br>8. Make sure the output follows the format specified in the constraint. |
| **prompt** | Transform the following Stack Overflow post into an Ansible playbook solution.<br>STACK OVERFLOW POST:<br>Title: title<br>Body: body<br>TARGET ENVIRONMENT:<br>Network:<br>- Subnet: 10.1.1.0/24<br>- Gateway: 10.1.1.254<br>Nodes:<br>1. ubuntu1 (10.1.1.1) - Ubuntu Linux<br>2. alpine1 (10.1.1.2) - Alpine Linux<br>3. centos1 (10.1.1.3) - CentOS Linux<br>4. redhat1 (10.1.1.4) - Red Hat Linux<br>IMPLEMENTATION CONSTRAINT: constraint<br>DELIVERABLE REQUIREMENTS:<br>1. Brief analysis of the problem and your solution approach<br>2. Production-ready, self-contained Ansible playbook that:<br><br>- Works across all specified distributions<br>- Includes all necessary variables and files<br>- Avoid using invalid file paths.<br>- Carefully examine the constraint and runtime environment, and select modules appropriate for both.<br>- Use only valid attributes for the selected modules.<br>- Make sure the output follows the format specified in the constraint.<br>- Follows Ansible's best practices<br><br>3. Must be enclosed in triple backticks for automated Python processing<br>Note: Focus on creating a robust, production-ready playbook that can be directly implemented in the target environment. |

Table 11: Level 3 TELeR (Santu and Feng, 2023) Prompt Structure with error awareness for IT Automation Tasks. Placeholder expressions enclosed in double curly braces (e.g., {{title}}, {{body}}, {{constraint}}) indicate variable components that will be dynamically replaced with task-specific information.

Table 12: Selected LLMs for Evaluation and Their Configuration Details. LLM Types indicate primary training focus (Code vs General). Sizes are approximate parameters.

| LLM Name | Abbreviation | Family | Size | Type | Download Date |
| --- | --- | --- | --- | --- | --- |
| WizardLMTeam/WizardCoder-15B-V1.0 (Luo et al., 2023) | WizardCoder-15B | WizardCoder | 15B | Code LLM | Jan 24 |
| codellama/CodeLlama-7b-Instruct-hf (Rozière et al., 2023) | CodeLlama-7B-it | CodeLlama | 7B | Code LLM | Apr 24 |
| codellama/CodeLlama-13b-Instruct-hf (Rozière et al., 2023) | CodeLlama-13B-it | CodeLlama | 13B | Code LLM | Apr 24 |
| bigcode/starcoder2-7b (Lozhkov et al., 2024) | Starcoder2-7B | StarCoder | 7B | Code LLM | Jul 24 |
| google/codegemma-7b-it (Zhao et al., 2024) | Codegemma-7B-it | CodeGemma | 7B | Code LLM | Aug 24 |
| Qwen/Qwen2.5-Coder-7B-Instruct (Hui et al., 2024) | Qwen2.5-Coder-7B-it | Qwen2.5 | 7B | Code LLM | Sept 24 |
| Qwen/Qwen2.5-7B-Instruct (Yang et al., 2024) | Qwen2.5-7B-it | Qwen2.5 | 7B | Code LLM | Sept 24 |
| lmsys/vicuna-7b-v1.5 (Zheng et al., 2023) | Vicuna-7B-it | Vicuna | 7B | General LLM | Mar 24 |
| meta-llama/Llama-3.1-8B-Instruct (Dubey et al., 2024) | Llama-3.1-8B-it | Llama 3.1 | 8B | General LLM | Oct 24 |
| meta-llama/Llama-3.2-3B-Instruct (Dubey et al., 2024) | Llama-3.2-3B-it | Llama 3.2 | 3B | General LLM | Oct 24 |
| microsoft/Phi-3.5-mini-instruct (Abdin et al., 2024) | Phi-3.5-mini-it | Phi 3.5 | 3.8B | General LLM | Sept 24 |
| deepseek-ai/DeepSeek-Coder-V2-Lite-Instruct (Zhu et al., 2024) | DeepSeek-Coder-V2-it | DeepSeek | 7B | Code LLM | Mar 25 |
| deepseek-ai/DeepSeek-R1-Distill-Llama-8B (DeepSeek-AI, 2025) | DeepSeek-Distil-L | DeepSeek | 8B | Reasoning LLM | Mar 25 |
| deepseek-ai/DeepSeek-R1-Distill-Qwen-7B (DeepSeek-AI, 2025) | DeepSeek-Distil-Q | DeepSeek | 7B | Reasoning LLM | Mar 25 |

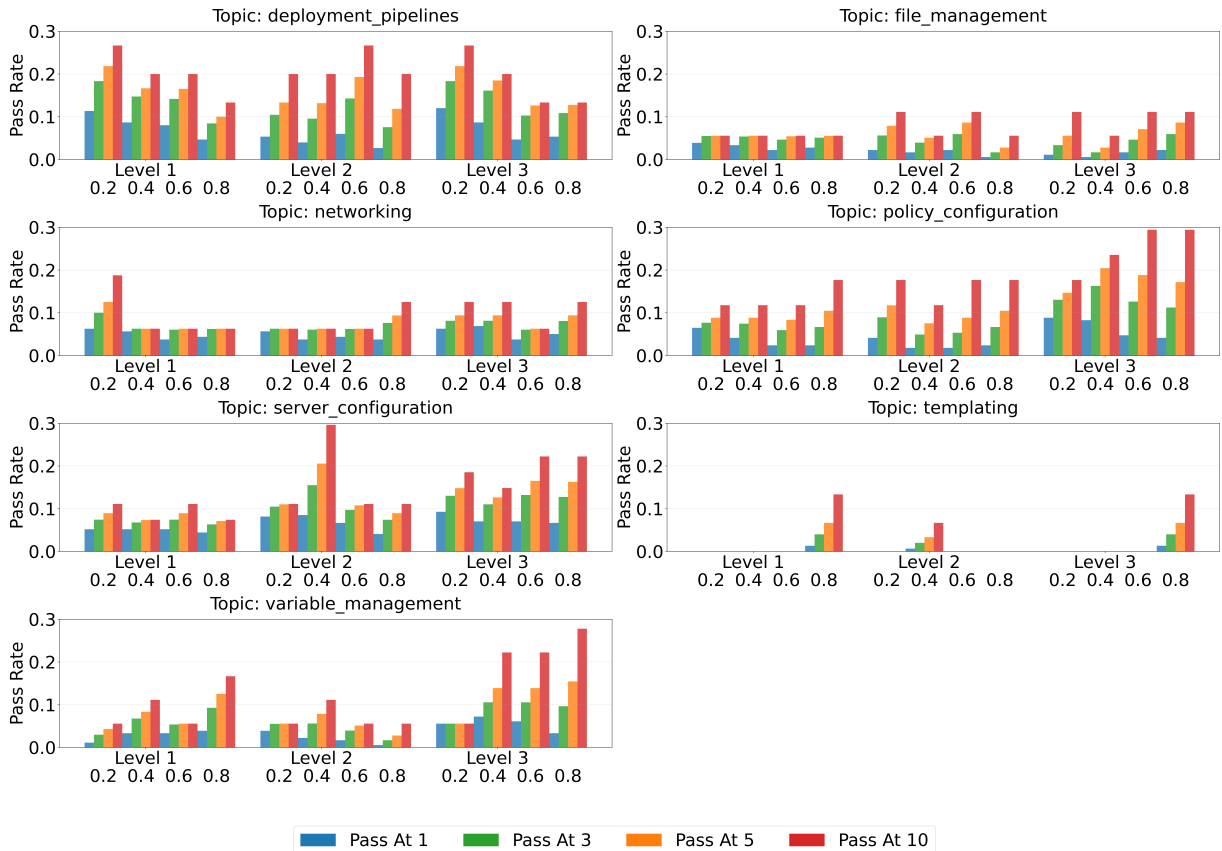

Figure 5: Performance of Qwen-2.5-Coder-7B model.

for the Pass@10 metric, while Pass@1 typically showed a downward trend. Specifically, when an upward trend was observed at higher temperatures, it often corresponded to a small number of successful generations (typically 1–3 out of 30 samples per issue) for highly customized tasks where memorized or common solutions likely failed. This suggests that higher temperatures, by generating a more diverse set of outputs and consequently a wider array of error messages, enabled the model to occasionally discover correct, albeit rare, solutions through increased exploration. Conversely, the downward trend in Pass@1 (i.e., better performance at lower temperatures) was more common for less customized problems, where the LLM might have already encountered similar code patterns, making deterministic generation more effective.

### D.1 Upward Trend

Here I have included the cases for which we see upward trend and their error analysis.

#### D.1.1 Task ID: 68411894

For this task, error trend can be found in Table 13.

Table 13: Error counts by temperature for Task ID 68411894.

| Error Name | T=0.2 | T=0.8 |
|---|---|---|
| Undefined Loop Variable | 20 | 10 |
| No Tasks | 2 | 0 |
| Wrong File Lookup | 1 | 0 |
| Wrong Path | 5 | 7 |
| Define Variable Itself & Fails | 1 | 0 |
| Invalid Code | 0 | 3 |
| Undefined Variable | 0 | 1 |
| Wrong Variable Access | 0 | 2 |
| Wrong Syntax for Block Used | 0 | 1 |
| Not Loaded Data | 0 | 1 |

#### D.1.2 Task ID: 44134617, Pass Count: 2

For Task ID 44134617, which involved defining server variables and then ensuring a specific directory structure and file creation (base directory /tmp/servers, individual server directories like /tmp/servers/server*, and a file within each), successful instances were observed with TELER Level 3 prompts (two passes). However, when examining failures at a sampling temperature of T=0.8, common issues included incorrect path usage (5 instances), such as attempting to use /tmp/server* directly where a loop or specific server name was required, and incorrect host targeting (1 instance).

### D.2 Task ID: 44738112, Pass Count: 1

Error diversity can be found in Table 14.

### D.3 Task ID: 51137660, Pass Count: 1

Table 15 summarizes the diversity of error types observed at different temperatures, highlighting how higher sampling temperature leads to a broader range of failure modes.

### D.4 Task ID: 60687917, Pass Count: 1

### D.5 Task ID: 28606876, Pass Count: 1

### D.6 Task ID: 39922427, Pass Count: 2

### D.7 Task ID: 46927378, Pass Count: 1

## E Detailed Unit Test Cases

### E.1 Server Configuration: User Check

```
---
- name: Check if user exists in /etc/
    passwd on all compute nodes
  hosts: all
  become: no
  vars:
    username: "{{ lookup('env', '
        username') }}"
  tasks:
    - name: Check if user exists in /etc
        /passwd
      command: grep "^{{ username }}:" /
          etc/passwd
      register: user_check
      ignore_errors: yes

    - name: Print result based on user
        existence
      debug:
        msg: "{{ username }} is present"
      when: user_check.rc == 0

    - name: Print result based on user
        absence
      debug:
        msg: "Not Found"
      when: user_check.rc != 0
```

Listing 2: User Check

**Expected Outcome:** The expected outcome involves the successful execution of the automation task, ensuring it completes without error for at least one designated user. Verification also includes confirming that the output accurately reports the number of specified items or users that were not found during the process. Furthermore, the system must correctly display or log the count of users who were successfully processed or identified in accordance with the task's specific objectives.

Table 14: Error diversity by temperature.

| Issue Description | Count (T=0.2) | Count (T=0.8) |
|---|---|---|
| ERROR! conflicting action statements | 2 | - |
| assert and fail message together | 1 | - |
| invalid code (syntactic/parsing errors) | 9 | 8 |
| invalid path (file/directory not found) | 17 | 12 |
| wrong inventory (host not in inventory) | 3 | - |
| wrong host (targeting incorrect host) | - | 6 |
| undefined variable | - | 1 |
| command not found (in shell module, etc.) | - | 1 |
| does not execute file (e.g., script module issue) | - | 2 |
| invalid operand used (e.g., in conditional) | - | 1 |

Table 15: Error diversity by temperature.

| Error Type | T = 0.2 | T = 0.8 |
|---|---|---|
| Undefined variable | 27 | 10 |
| Wrong attribute accessed | 6 | present |
| Wrong function applied | 1 | – |
| Syntax issue | – | 13 |
| Invalid attribute used | – | present |
| Invalid YAML | – | 3 |
| Does not filter non-running services | – | 2 |
| Could not filter out services | – | 1 |
| Unsupported parameter used | – | 1 |
| Invalid operand used | – | 1 |

### E.2 Policy Configuration: File existance check

```yaml
---
- name: WAMUI Prerequisites Installer
  hosts: localhost
  gather_facts: no
  tasks:
    - name: Ensure required files exist
      copy:
        src: data/metadata/44738112/file
            .bin
        dest: /tmp/file.bin
        mode: '0755'

    - name: Run file.bin with expect
        module
      expect:
        command: '/tmp/file.bin'
        responses:
          'PRESS <ENTER> TO CONTINUE:':
              '\n'
      register: expect_result

    - name: Display expect module output
      debug:
        msg: "{{ expect_result.
            stdout_lines }}"
```

Listing 3: File existance check.

**Expected Outcome:** The expected outcome for this task is the successful execution of the playbook on the controller node. This entails utilizing the specified file, 'data/metadata/44738112/file.bin', located on the controller. Additionally, verification includes ensuring that all intended output is correctly printed to localhost and confirming that the playbook operated on the correct inventory, targeting the controller node as specified.

### E.3 Networking: Configure rax

```yaml
---
- name: Conditional Group Variable
    Example
  hosts: localhost
  connection: local
  gather_facts: no

  vars:
    rax_count: '2'
    rax_group: 'web_group'

  tasks:
    - name: Check if rax_count is
        greater than 1
      set_fact:
```

Table 16: Error diversity for Task ID 60687917 by temperature.

| Error Type | T = 0.2 | T = 0.8 |
| --- | --- | --- |
| Type mismatch in comparison | 15 | 15 |
| Invalid YAML | 13 | 9 |
| Wrong host condition | 3 | 3 |
| Wrongly setting var to default | 2 | – |
| Conflicting action statement | 5 | 1 |
| Wrong host | 1 | 4 |
| Undefined variable | – | 1 |
| Wrong attribute for block | – | 1 |
| Template error | – | 1 |
| Hosts field not set | – | 2 |

Table 17: Error diversity for Task ID 28606876 by temperature.

| Error Type | T = 0.2 | T = 0.8 |
| --- | --- | --- |
| Cannot resolve socket | 3 | 3 |
| Invalid YAML | 14 | 9 |
| lsof not found in shell | 1 | 1 |
| Undefined attribute used | 5 | – |
| Undefined variable used | 2 | – |
| Timeout | 1 | 1 |
| Wrong service checked | 2 | 1 |
| Malformed data passed | 2 | – |
| Invalid value in hosts | – | 1 |
| Unsupported parameter used | – | 2 |
| Wrong attribute used | – | 4 |
| Invalid variables specified | – | 1 |
| Loop inside block is invalid | – | 1 |
| Unbalanced Jinja block | – | 1 |

```yaml
          rax_group: 'web_group'
      when: rax_count | int > 1

    - name: Debug rax_count
      debug:
        msg: "rax_count is {{ rax_count
            }}"

    - name: Debug rax_group
      debug:
        msg: "rax_group is {{ rax_group
            }}"
```

Listing 4: Ansible Playbook for rax Configuration

**Expected Outcome:** Check if rax group has correct count..

### E.4 Deployment Pipelines: Server existence check and write on file

```yaml
---
- name: Check service existence and
    write to file
  hosts: all
  become: yes
  tasks:
    - name: Check if dummy_service.txt
        exists
      stat:
        path: /tmp/63688612/
            dummy_service.txt
      register: service_check

    - name: Write "The Service Exists"
        to test.txt if file exists
      copy:
        content: "The Service Exists"
        dest: /tmp/63688612/test.txt
      when: service_check.stat.exists

    - name: Write "The Service Does Not
        Exist" to test.txt if file does
        not exist
      copy:
        content: "The Service Does Not
            Exist"
        dest: /tmp/63688612/test.txt
```

Table 18: Error diversity for Task ID 39922427 by temperature.

| Error Type | T = 0.2 | T = 0.8 |
| --- | --- | --- |
| Wrong inventory | 8 | 4 |
| Invalid variable | 9 | – |
| Invalid path | 8 | – |
| Undefined path | – | 1 |
| Undefined variable | – | 2 |
| Connection refused | – | 1 |
| Syntax issue | – | 1 |
| *Check mode usage:* | | |
| Does not use check mode | present | – |
| Uses check mode | – | present |

Table 19: Error diversity for Task ID 46927378 by temperature.

| Error Type | T = 0.2 | T = 0.8 |
| --- | --- | --- |
| Wrong URL | 20 | 14 |
| Invalid path | 7 | 2 |
| Invalid attribute for Ansible | 6 | – |
| Undefined variable | 10 | 8 |
| Format error | 1 | 1 |
| Wrong host | 2 | – |
| Timeout | 1 | – |
| Invalid YAML | 1 | 4 |
| No module/action in task | – | 5 |
| Unnecessary package install | – | 2 |
| Invalid role used | – | 1 |
| Wrong attribute for block | – | 1 |
| Wrong module used | – | 3 |
| Missing argument | – | 1 |

```
      when: not service_check.stat.
          exists
```

Listing 5: Ansible Playbook for Server existence check and write on file

**Expected Outcome:** The task involves performing a file-based service existence check. Specifically, the automation script must determine if a file exists at the path /tmp/63688612/dummy_service.txt. Based on this check, it must write either 'The Service Exists' or 'The Service Does Not Exists' into the file /tmp/63688612/test.txt.

The verification process occurs in two stages. First, with the dummy_service.txt file absent, the playbook is expected to pass execution on all nodes. Success in this stage is further confirmed by verifying that the test.txt file is created on all nodes and that its content correctly states 'The Service Does Not Exists'. Subsequently, the dummy_service.txt file is introduced into the environment. The playbook is then executed again, and it is expected to pass on all nodes. Verification for this second stage involves checking that the test.txt file is present on all nodes and, crucially, that its contents now correctly reflect 'The Service Exists'.

### E.5 Variable Management: Transform Directory Substitution

```
---
- name: Transform People Dictionary
  hosts: localhost
  gather_facts: false

  vars_files:
    - "data/metadata/35605603/result.yml
        "
```

```yaml
  tasks:
    - name: Initialize the transformed
        dictionary
      set_fact:
        people_dict: {}

    - name: Populate the dictionary with
         names and genders
      set_fact:
        people_dict: "{{ people_dict |
            combine({ item.item.name:
            item.stdout }) }}"
      loop: "{{ people.results }}"

    - name: Print the transformed
        dictionary
      debug:
        var: people_dict
```

Listing 6: Ansible Playbook for Transforming directory

**Expected Outcome:** The task requires utilizing the people variable, which is to be sourced from the file 'data/metadata/35605603/result.yml'. The playbook's primary action is to print this variable after it has been converted into a dictionary format. This operation is to be performed specifically on the controller node.

Verification involves several checks: first, the playbook must be executed. Its successful completion without errors is then assessed. Crucially, the output must be inspected to ensure that the correct dictionary values derived from the people variable are printed. Finally, it is verified that the playbook ran on the intended inventory, targeting only the controller node.

### E.6 Templating: Check DNS reverse File

```yaml
---
- name: Check for DNS reverse files on
    compute nodes
  hosts: all
  become: yes

  vars:
    dns_reverse_dir: "/tmp/test_dns"
    dns_reverse_pattern: "^named\\."

  tasks:
    - name: Find DNS reverse files
      find:
        paths: "{{ dns_reverse_dir }}"
        patterns: "{{
            dns_reverse_pattern }}"
        use_regex: yes
      register: find_results

    - name: Print found DNS reverse
        files
      debug:
        msg: "Node {{ inventory_hostname
            }}: Found file - {{ item.
            path }}"
      loop: "{{ find_results.files }}"
      when: find_results.files | length
          > 0

    - name: Handle case when no DNS
        reverse files are found
      debug:
        msg: "Node {{ inventory_hostname
            }}: No DNS reverse files
            found."
      when: find_results.files | length
          == 0
```

Listing 7: Ansible Playbook for Checking DNS reverse file

**Expected Outcome:** The task requires the automation script to iterate through each compute node, identify any DNS reverse files located within the /tmp/test_dns directory, and print their names. The output for each found file should adhere to the format: "Node <node_name>: Found file - <file_path>". A critical aspect of this task is the graceful handling of errors, particularly if the /tmp/test_dns directory or any expected files do not exist.

Verification of the generated solution is multifaceted. Initially, the correctness of debug arguments used in the playbook is checked. The playbook is then tested in a scenario where the /tmp/test_dns directory is absent, ensuring that this condition is handled gracefully on each node. Subsequently, the directory is created, and the playbook's behavior with an empty directory is verified on all nodes. The testing progresses by partially populating the directory with some DNS reverse files; the solution must correctly handle this partial existence on each node. Throughout these stages, a key validation point is that the script accurately prints the names of any existing files in the specified format.

### E.7 File Management: Copy file to destination

```yaml
---
- name: Copy files to destination
    directory
  hosts: all
  become: yes
  vars:
    file_path: "/tmp/destination_dir"

  tasks:
    - name: Create destination directory
        if it does not exist
      file:
        path: "{{ file_path }}"
        state: directory
        mode: '0755'

    - name: Copy files to destination
        directory
```

```yaml
      copy:
        src: "data/metadata/{{ item }}"
        dest: "{{ file_path }}/{{ item
           }}"
        mode: '0755'
      loop:
        - "file1.txt"
        - "file2.txt"
```

Listing 8: Ansible Playbook for copy file to destination

**Expected Outcome:** Directory and file exists in all nodes.

# F Results

## F.1 Overall Performance Across Models

Overall Performance of all models are presented in Figure 5 - 17.

- Present main pass@k results averaged across all tasks.
- Compare performance across the 14 LLMs.
- Include results related to temperature effects.
- Include results related to prompt style effects.

## F.2 Performance by IT Automation Domains

To understand how model performance varies across different types of IT automation tasks, we analyzed the results within the seven functional categories derived from Stack Overflow (Section 3.1). Table 20 summarizes the performance, showing the number of tasks within each category and identifying the best-performing models, other models that successfully solved at least one task, and models that failed to solve any tasks in that category. A task was considered "solved" by a model if it produced at least one functionally correct and idempotent solution (pass@15 > 0, see Section C.3.1) across all tested prompt and temperature configurations.

The results reveal significant variation in performance across both models and task categories. Key observations include:

- **Dominant Performance:** The specialized code model `Qwen/Qwen2.5-Coder-7B-Instruct` demonstrated exceptionally strong performance, solving
- **Strong Generalists/Code Models:** `meta-llama/Llama-3.1-8B-Instruct` and `google/codegemma-7b-it` also performed well overall, showing particular strength in *Server Configuration* and achieving perfect scores in *File Management*. Other models like `bigcode/starcoder2-7b` and `codellama/CodeLlama-13b-Instruct-hf` were also competitive in *Server Configuration*.
- **Category Difficulty:** *Variable Management* and *Templating* appear to be the most challenging categories for the evaluated LLMs. `Qwen/Qwen2.5-Coder-7B-Instruct` was the only model to solve all Variable Management tasks, and many models failed to solve any tasks in these categories (see Table 20, last column). This suggests these tasks require deeper understanding of Ansible's execution logic, variable precedence, and Jinja2 templating specifics. *Networking* and *Policy Configuration* also presented challenges for several models.
- **Easier Categories:** *Server Configuration* and *File Management* saw broader success across more models, likely reflecting the commonality of these tasks and potentially simpler state verification required. Notably, no model failed to solve at least some *Server Configuration* tasks.
- **Limited Performance Models:** Certain models, particularly [DS-R1-Llama] and [DS-R1-Qwen] (assuming these are distinct models), consistently failed across multiple categories, indicating potential limitations in their training or fine-tuning for IaC tasks.

This category-level analysis highlights specific strengths and weaknesses of current LLMs for IaC generation. The variance suggests that while some tasks are within reach, complex configuration logic, particularly involving templating and variable interactions, remains a significant hurdle. These findings inform our subsequent qualitative analysis of error patterns (Section 5) and the investigation into the effects of prompting and temperature (Section 7).

# G Error Taxonomy

## G.1 Overall Trend

As we saw the low pass@k values, we wanted to investigate what really lies behind the numbers. So we did an extensive qualitative study where we studied 1411 failed cases spanning across task categories and models. In the process we have created a taxonomy of error in opensource models for ansible code when solving user-generated real world issues.

Our error taxonomy spans nine categories that capture the full spectrum of failures in Ansible playbook generation. Most strikingly, reasoning-

Table 20: Summary of model performance across the seven IaC task categories. '# Tasks' indicates the number of benchmark tasks in each category. Models listed solved at least one task unless noted in the last column. Counts in parentheses indicate the number of tasks solved by that specific model within the category. Assumes 'solved' means pass@1 > 0 for at least one configuration.

| Category | # Tasks | Best Model(s) (Solved) | Other Successful Models (Solved) | Models with No Success (Solved=0) |
|---|---|---|---|---|
| Variable Management | 8 | Qwen/Qwen2.5-Coder-7B-Instruct (8) | meta-llama/Llama-3.1-8B-Instruct (3), microsoft/Phi-3.5-mini-instruct (2), lmsys/vicuna-7b-v1.5 (2), WizardLMTeam/WizardCoder-15B-V1.0 (1), google/codegemma-7b-it (1) | codellama/CodeLlama-13b-Instruct-hf, codellama/CodeLlama-7B-Instruct-hf, Qwen/Qwen2.5-7B-Instruct, meta-llama/Llama-3.2-3B-Instruct, bigcode/starcoder2-7b, [DS-R1-Llama], [DS-R1-Qwen] |
| Networking | 4 | Qwen/Qwen2.5-Coder-7B-Instruct (4) | Qwen/Qwen2.5-7B-Instruct (3), bigcode/starcoder2-7b (3), codellama/CodeLlama-7B-Instruct-hf (2), meta-llama/Llama-3.1-8B-Instruct (2), google/codegemma-7b-it (2), microsoft/Phi-3.5-mini-instruct (1), lmsys/vicuna-7b-v1.5 (1), codellama/CodeLlama-13b-Instruct-hf (1) | [DS-R1-Qwen] |
| Templating | 4 | Qwen/Qwen2.5-Coder-7B-Instruct (3) | google/codegemma-7b-it (2), microsoft/Phi-3.5-mini-instruct (2), codellama/CodeLlama-7B-Instruct-hf (1), codellama/CodeLlama-13b-Instruct-hf (1), Qwen/Qwen2.5-7B-Instruct (1), meta-llama/Llama-3.1-8B-Instruct (1) | WizardLMTeam/WizardCoder-15B-V1.0, bigcode/starcoder2-7b, lmsys/vicuna-7b-v1.5, meta-llama/Llama-3.2-3B-Instruct, [DS-R1-Llama], [DS-R1-Qwen] |
| Deployment Pipelines | 4 | Qwen/Qwen2.5-Coder-7B-Instruct (4), meta-llama/Llama-3.1-8B-Instruct (4) | codellama/CodeLlama-13b-Instruct-hf (3), google/codegemma-7b-it (2), microsoft/Phi-3.5-mini-instruct (2), lmsys/vicuna-7b-v1.5 (2), bigcode/starcoder2-7b (2), Qwen/Qwen2.5-7B-Instruct (1), WizardLMTeam/WizardCoder-15B-V1.0 (1) | [DS-R1-Llama], [DS-R1-Qwen] |
| Policy Configuration | 7 | Qwen/Qwen2.5-Coder-7B-Instruct (7) | lmsys/vicuna-7b-v1.5 (4), microsoft/Phi-3.5-mini-instruct (2), google/codegemma-7b-it (2), WizardLMTeam/WizardCoder-15B-V1.0 (1), bigcode/starcoder2-7b (1), meta-llama/Llama-3.1-8B-Instruct (1), [DS-R1-Llama] (2), meta-llama/Llama-3.2-3B-Instruct (1) | codellama/CodeLlama-13b-Instruct-hf, codellama/CodeLlama-7B-Instruct-hf, Qwen/Qwen2.5-7B-Instruct |
| Server Configuration | 11 | Qwen/Qwen2.5-Coder-7B-Instruct (11) | meta-llama/Llama-3.1-8B-Instruct (10), bigcode/starcoder2-7b (9), google/codegemma-7b-it (8), codellama/CodeLlama-13b-Instruct-hf (8), Qwen/Qwen2.5-7B-Instruct (8), codellama/CodeLlama-7B-Instruct-hf (7), lmsys/vicuna-7b-v1.5 (5), meta-llama/Llama-3.2-3B-Instruct (5), microsoft/Phi-3.5-mini-instruct (4), WizardLMTeam/WizardCoder-15B-V1.0 (3) | — |
| File Management | 5 | google/codegemma-7b-it (5), meta-llama/Llama-3.1-8B-Instruct (5) | Qwen/Qwen2.5-Coder-7B-Instruct (4), bigcode/starcoder2-7b (3), codellama/CodeLlama-13b-Instruct-hf (3), WizardLMTeam/WizardCoder-15B-V1.0 (2), lmsys/vicuna-7b-v1.5 (2), Qwen/Qwen2.5-7B-Instruct (2), microsoft/Phi-3.5-mini-instruct (2), meta-llama/Llama-3.2-3B-Instruct (2), codellama/CodeLlama-7B-Instruct-hf (1) | [DS-R1-Llama], [DS-R1-Qwen] |

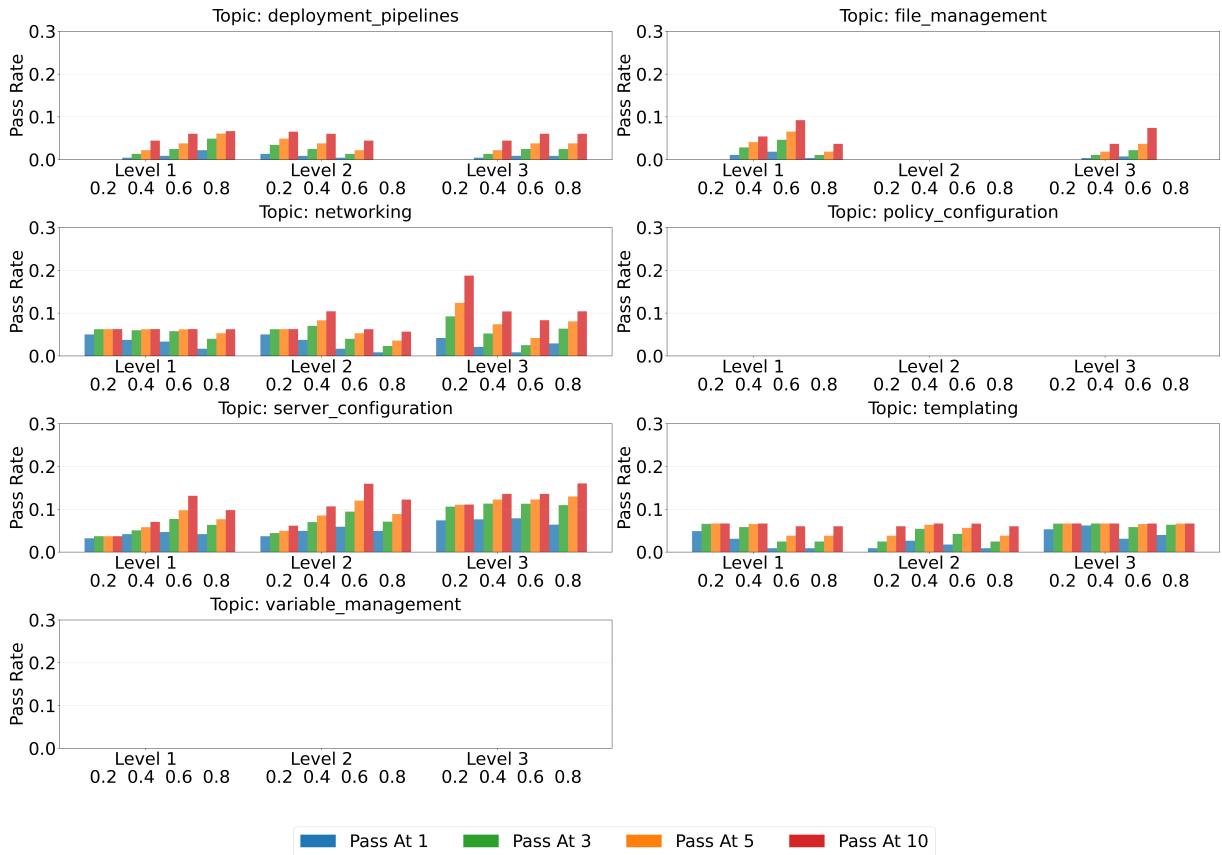

Figure 6: Performance of Qwen-2.5-7B-instruct-it model.

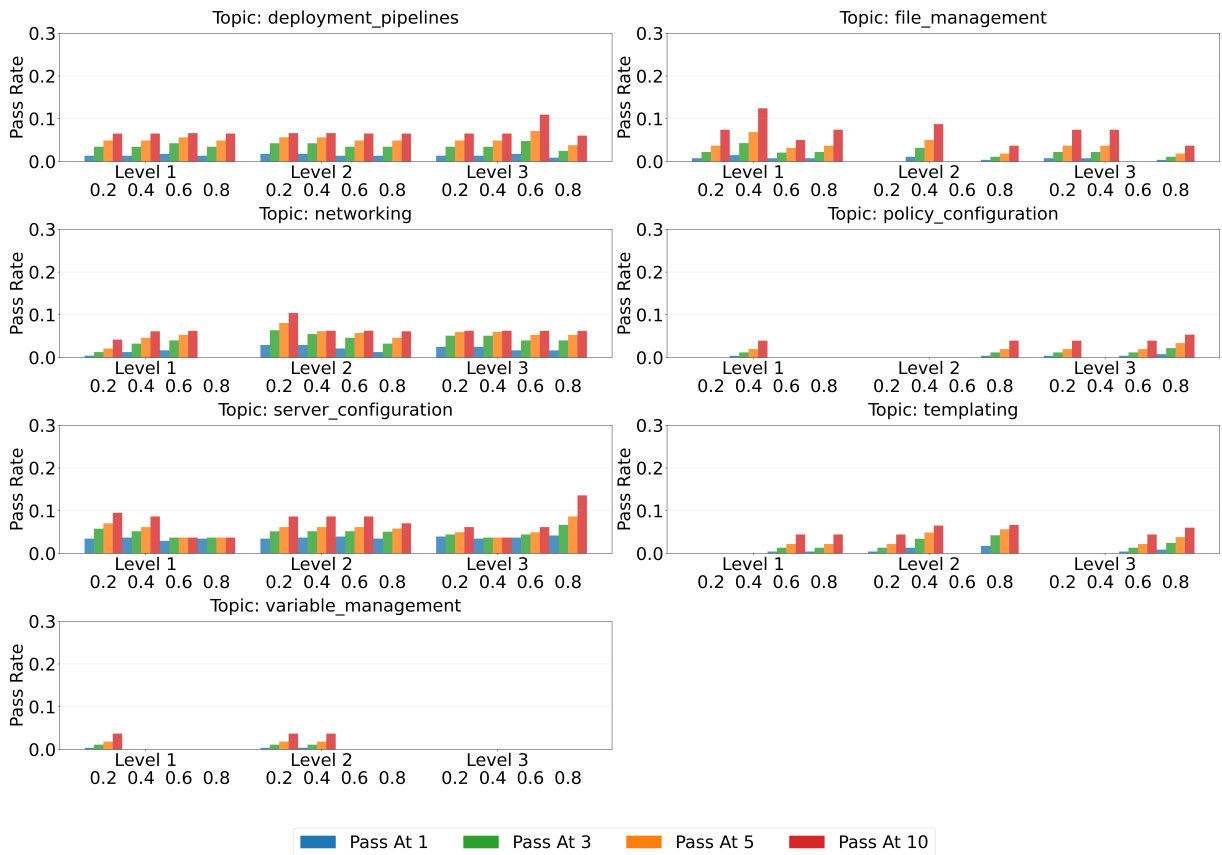

Figure 7: Performance of CodeGemma-7B-it model.

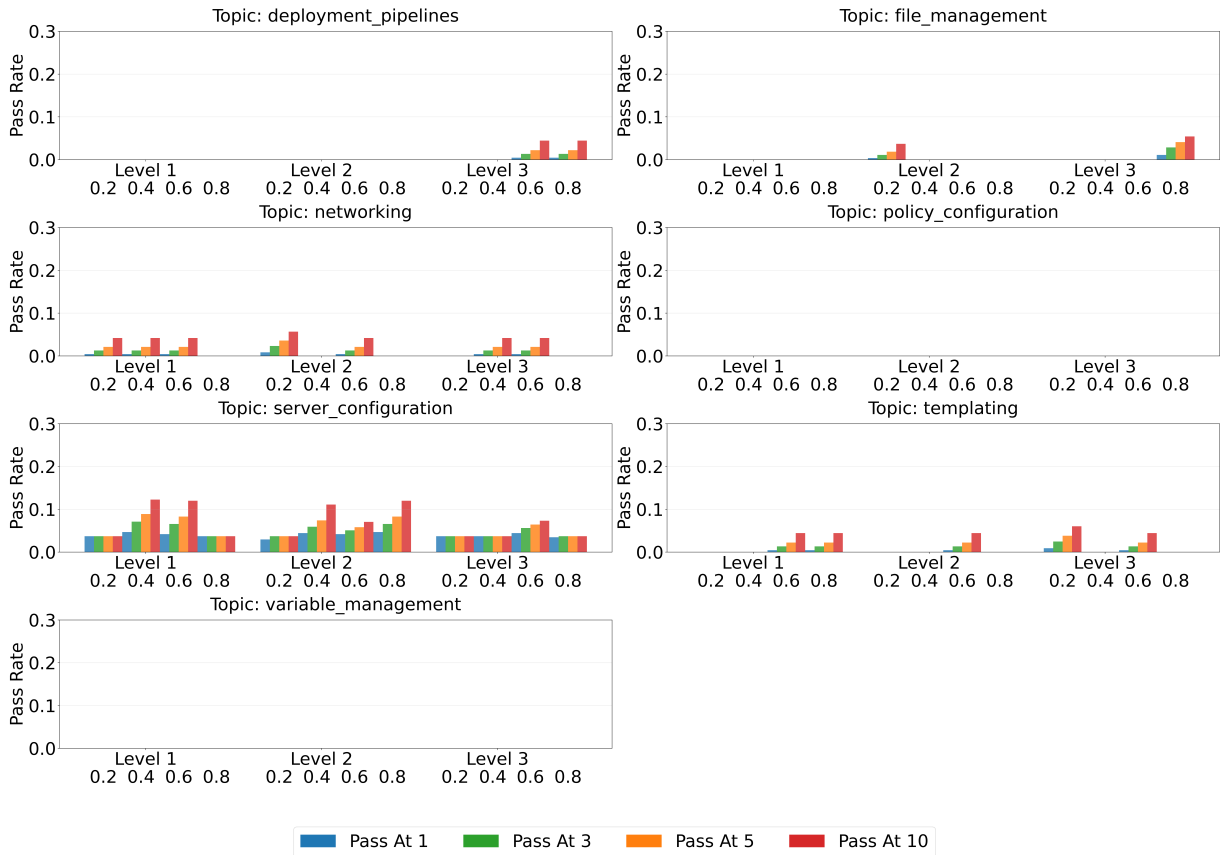

Figure 8: Performance of CodeLLaMa-7B-it model.

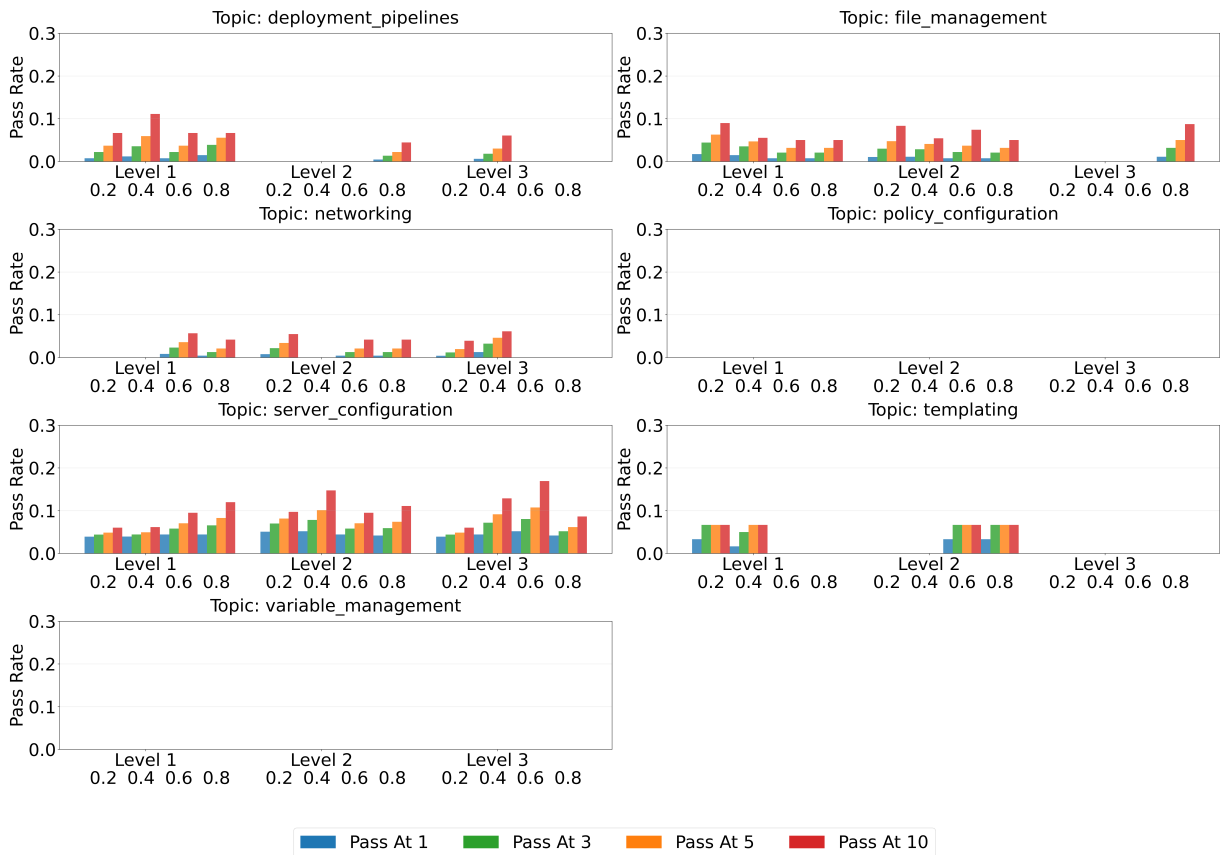

Figure 9: Performance of CodeLLaMa-13B-it model.

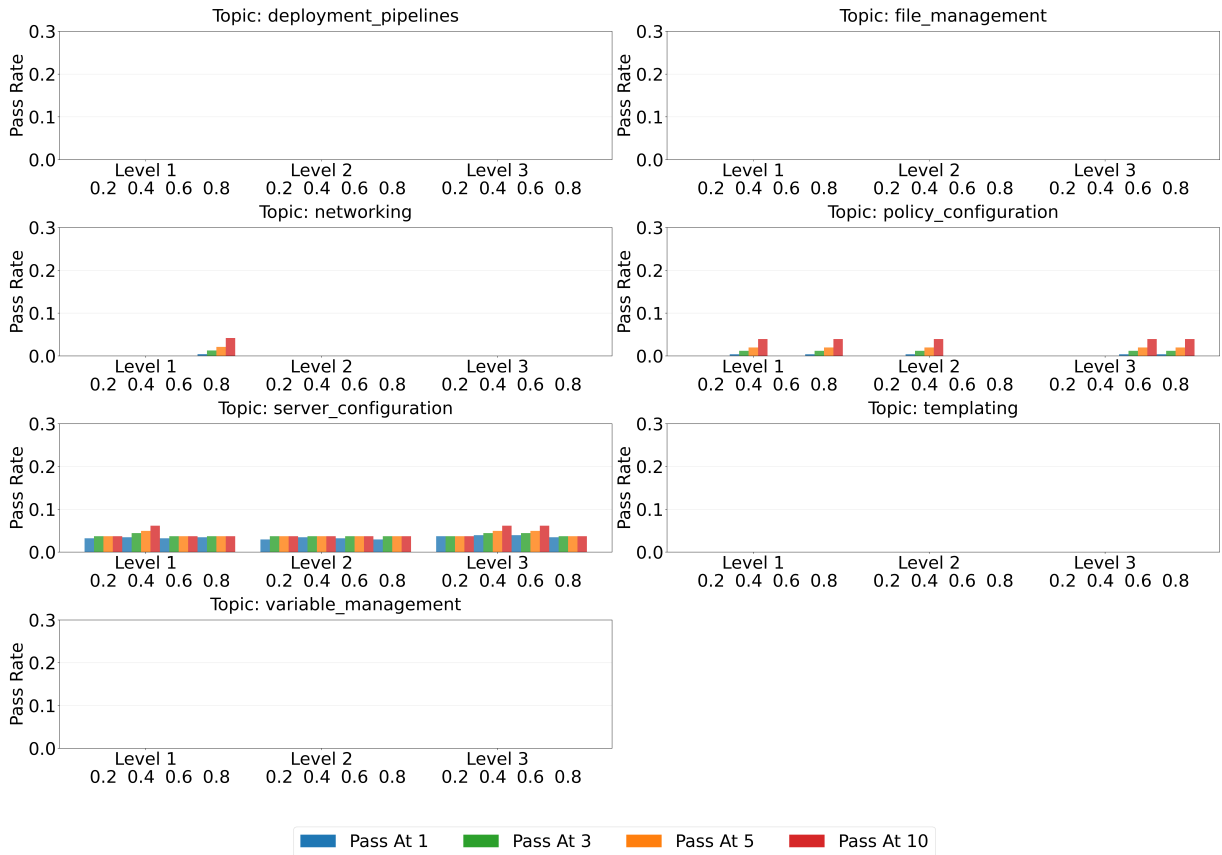

Figure 10: Performance of DeepSeek-R1-Distil-LLaMa-8B model.

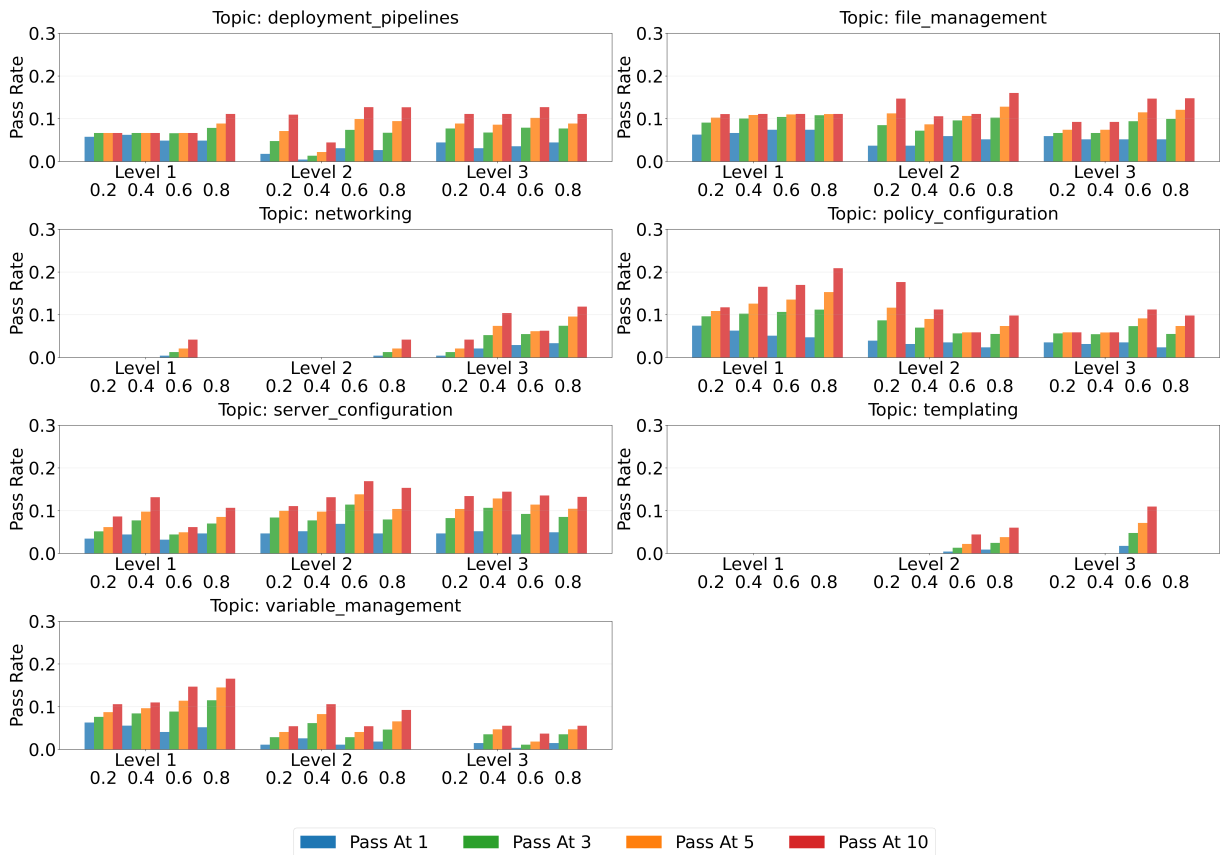

Figure 11: Performance of DeepSeek-Coder-V2-it model.

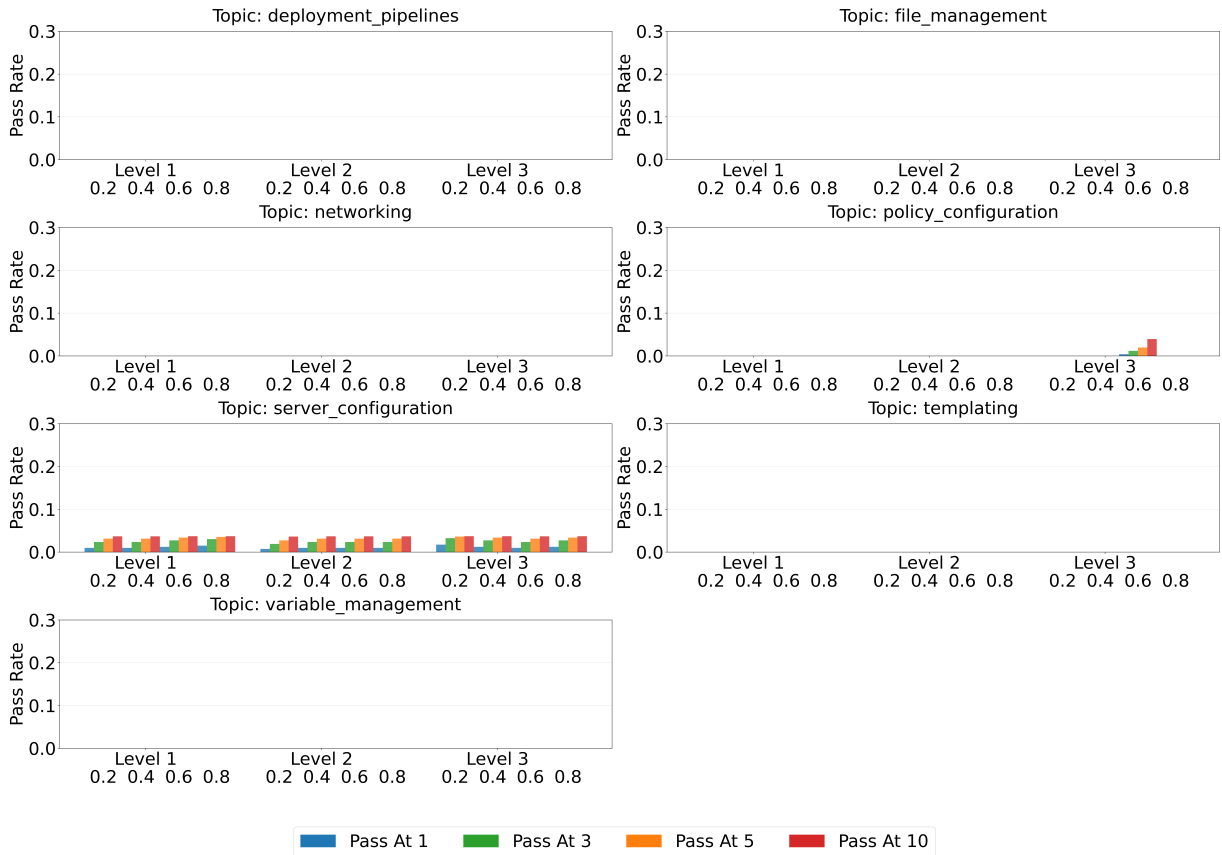

Figure 12: Performance of DeepSeek-R1-Distil-Qwen-7B model.

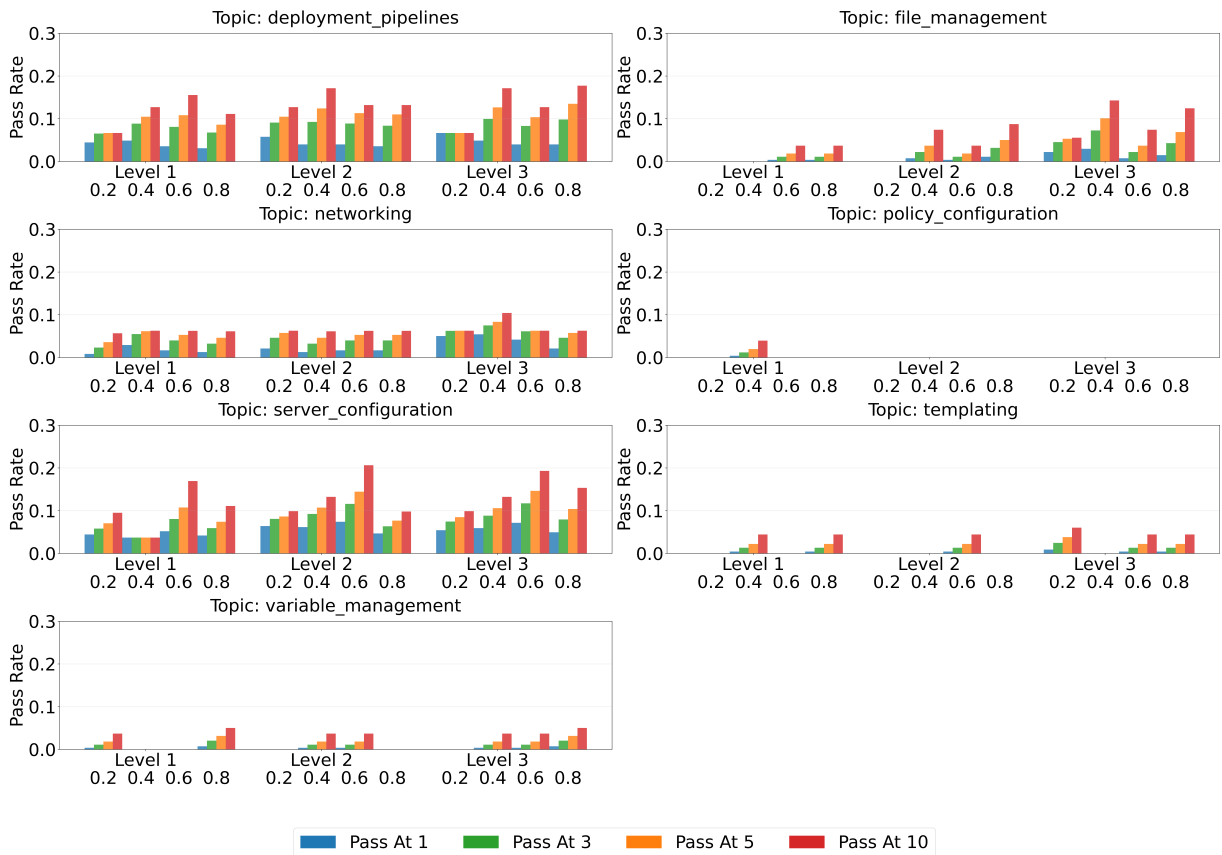

Figure 13: Performance of LLaMa-3.1-8B-it model.

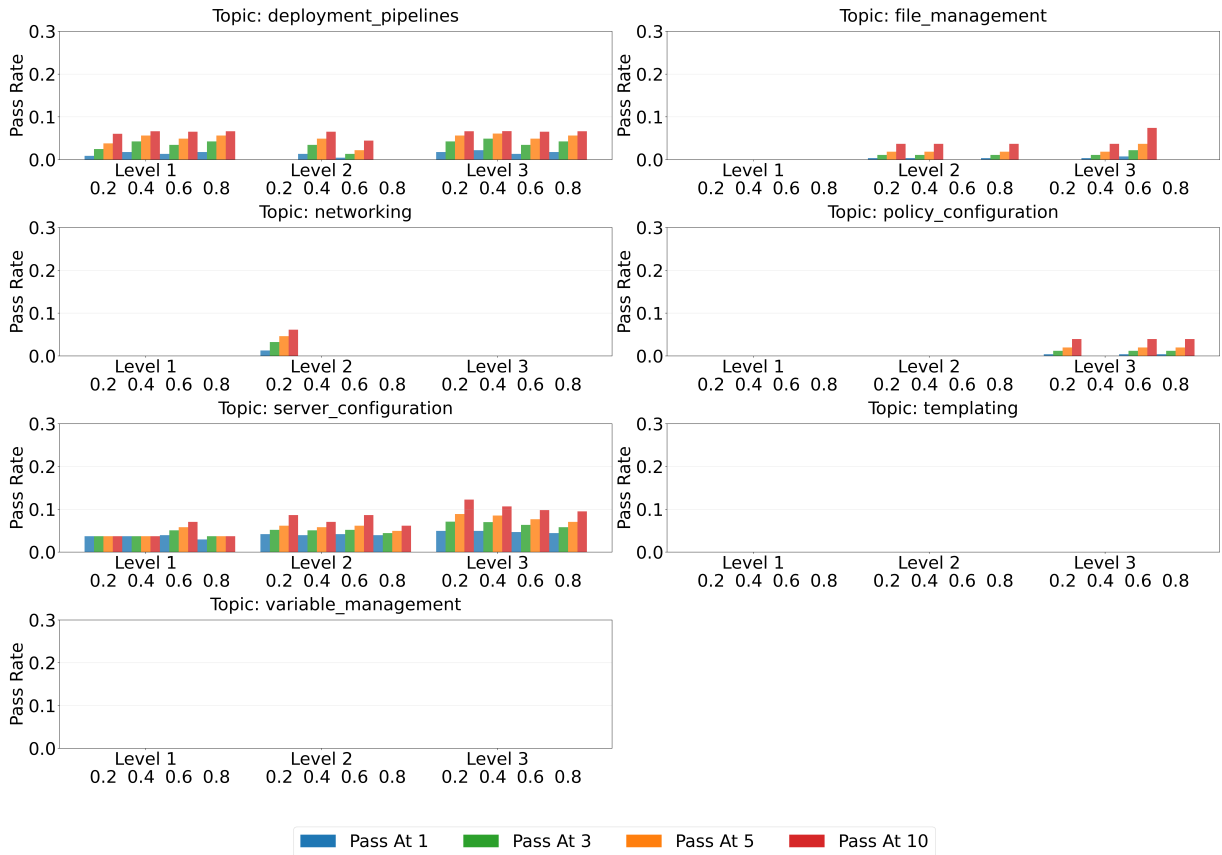

Figure 14: Performance of LLaMa-3.2-3B-it model.

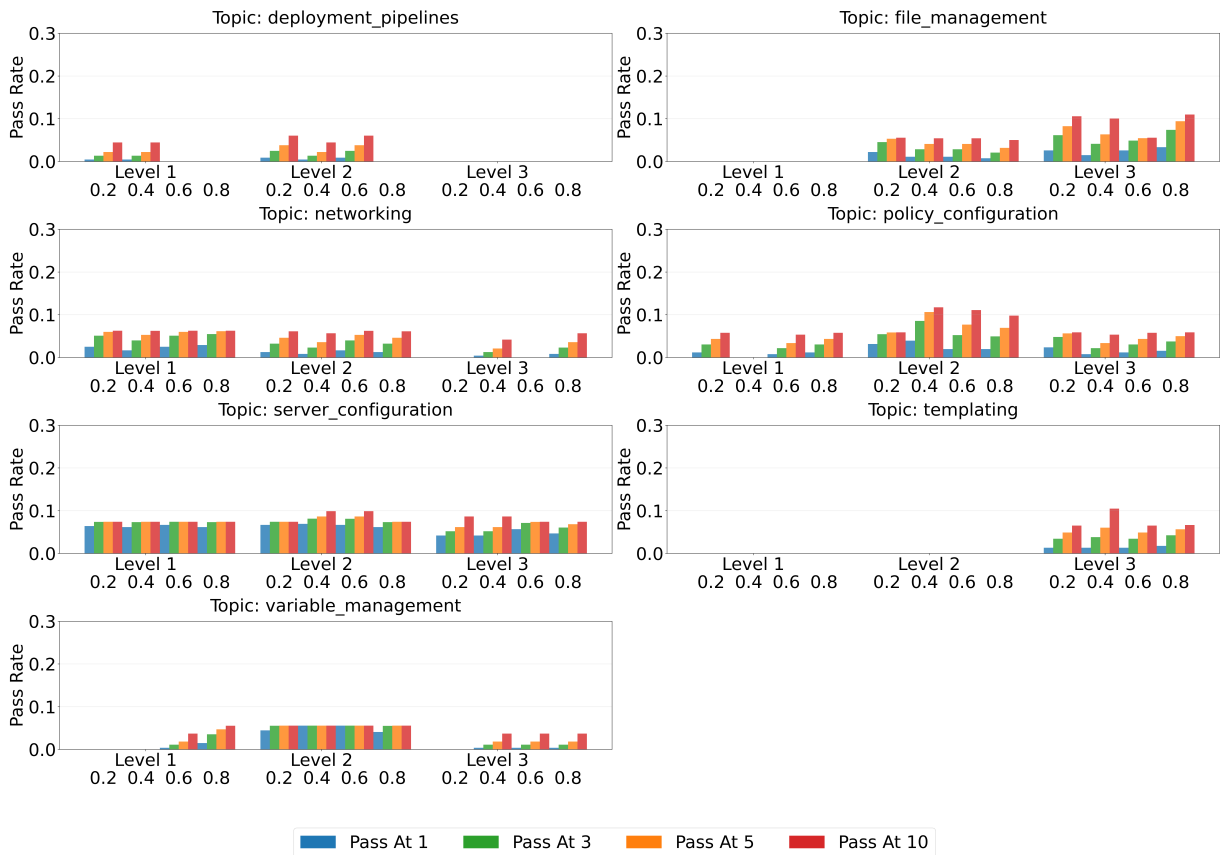

Figure 15: Performance of Microsoft-phi-3.5-Mini-it model.

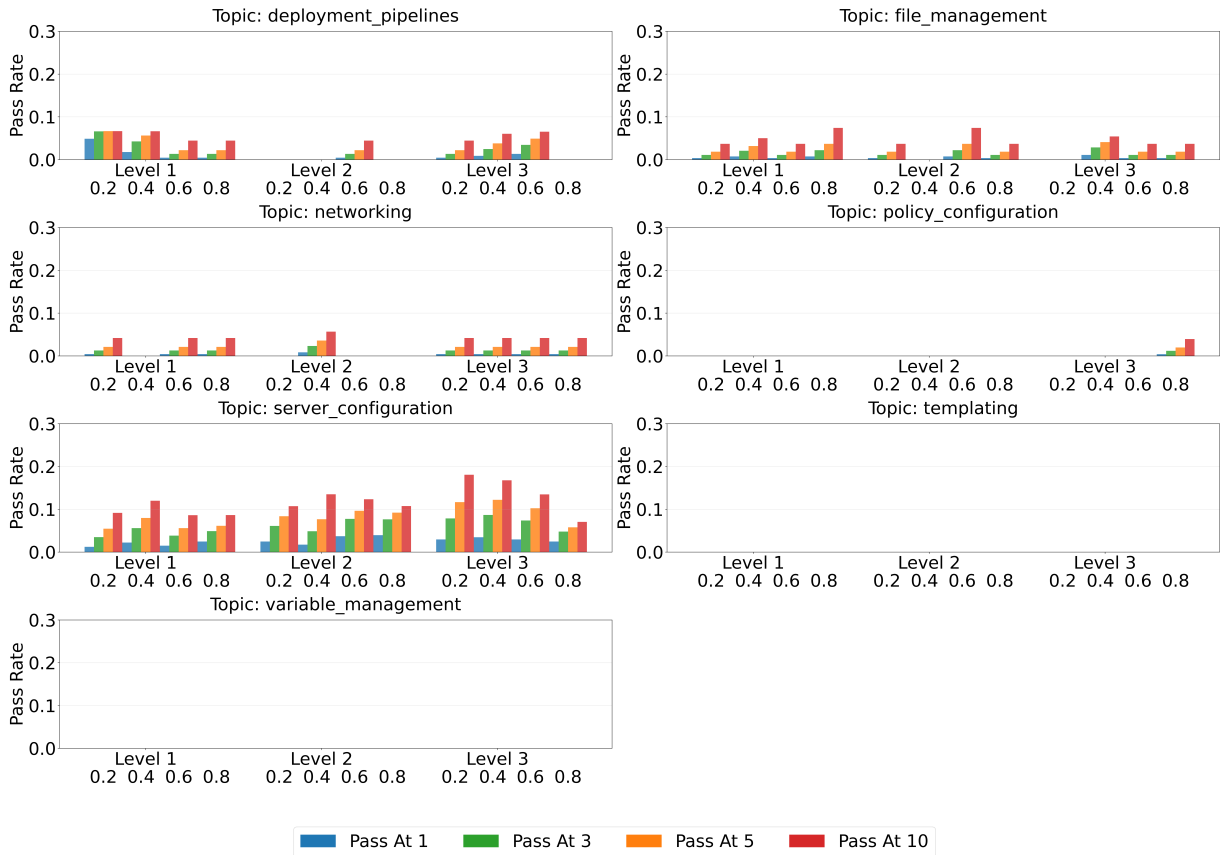

Figure 16: Performance of StarCoder2-7B model.

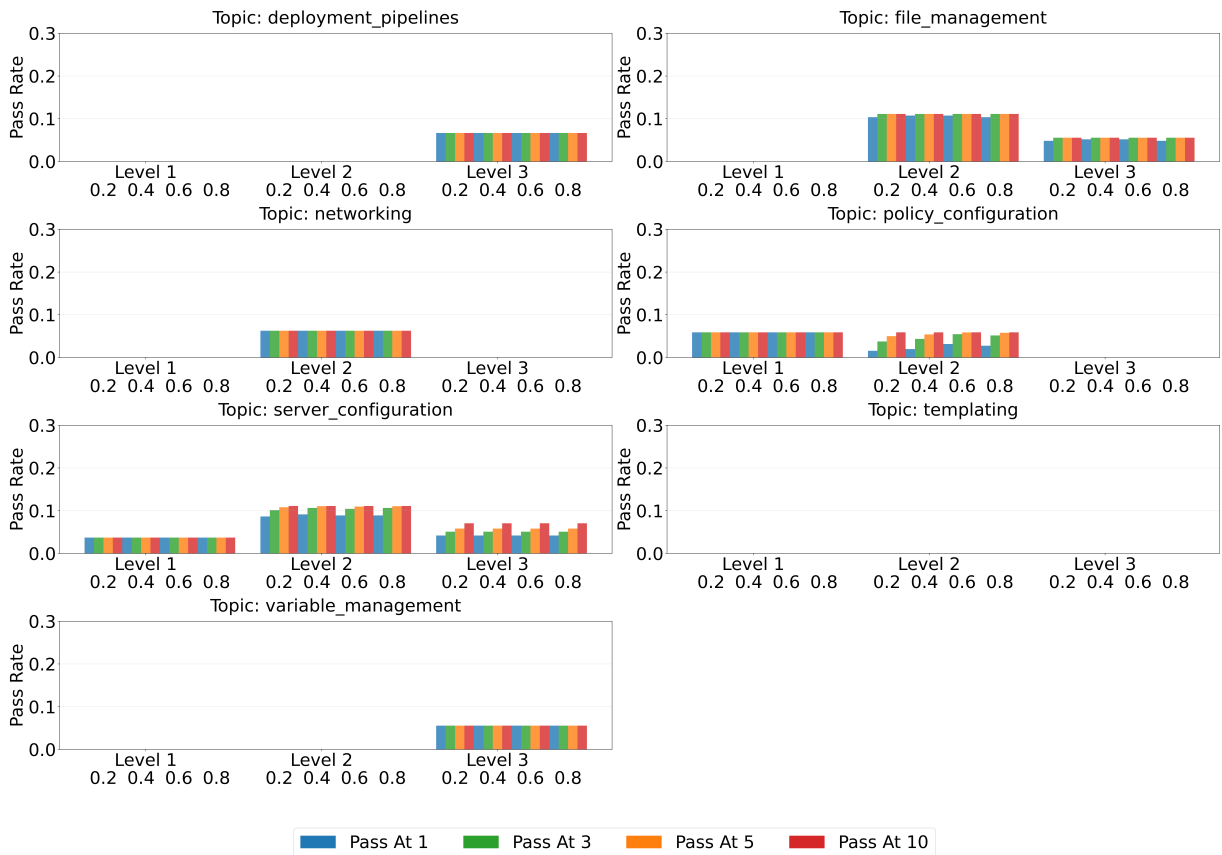

Figure 17: Performance of WizardCoder-15B model.

distilled models exhibit a fundamentally different failure profile than standard code generation models-despite their enhanced reasoning capabilities, they overwhelmingly fail at basic syntax, suggesting that general reasoning abilities do not transfer to domain-specific code generation without appropriate knowledge of the target language.

For standard code generation models, the error distribution reveals more nuanced challenges. Attribute & Parameter Errors emerge as the most prevalent semantic failure mode, where models correctly identify appropriate modules but fail to configure them properly. This suggests that models understand "what" to do but struggle with "how" to do it correctly-a crucial distinction from traditional code generation tasks that often focus on algorithmic logic rather than precise configuration.

Variable handling and path navigation errors collectively account for a substantial portion of failures, highlighting the challenge of state tracking in IaC. Unlike traditional programming where variable scope is often contained within functions, Ansible's variable handling spans playbooks, roles, and templates, creating a more complex state management problem that current models struggle to navigate.

Template issues, particularly with Jinja2, reveal a fundamental limitation in models' ability to reason about the interaction between static configuration and dynamic content generation-a core requirement for flexible infrastructure automation. This challenge is compounded by the dual-language nature of templating, where models must simultaneously reason about Ansible's declarative structure and Jinja2's programming constructs.

The distribution of errors across models also reveals architectural influences: some models struggle predominantly with variable issues, while others face challenges with host targeting or attribute configuration. These patterns suggest that different pre-training approaches and model architectures may create distinct blindspots in handling IaC's unique requirements.

To better understand the failure modes of LLM-generated Ansible playbooks, we performed a systematic error analysis for each IaC category. For each category, we selected a representative task and, for each model, sampled 15 failed playbooks stratified by temperature and TELeR prompt level. Each sampled playbook was manually reviewed and categorized by error type. The following subsections and tables summarize the most frequent error patterns observed for each model in each category.

### G.1.1 Networking: Task ID 24269533

This networking task involves defining two variables and printing them, making it relatively straightforward. As expected, the pass rate across models was relatively high. However, common error types still emerged across models, often related to basic syntax, variable definition, and adherence to constraints.

Many models, such as CodeGemma and CodeLLaMa variants, failed by referencing undefined or out-of-scope variables. Several models (e.g., CodeLLaMa, LLaMA-3.2-3B, Phi-3.5, WizardCoder) included tasks that were explicitly prohibited in the prompt, such as provisioning a web server, indicating prompt misunderstanding or prompt misalignment. Invalid host configuration and YAML syntax issues also appeared frequently (notably in Qwen-Instruct and StarCoder2), which hindered execution even when the task logic was close to correct. A few models, such as DeepSeek-R1-Distil-Qwen, provided inconsistent or incomplete playbook structures without producing valid YAML. Overall, the results highlight that even in simple tasks, LLMs often struggle with instruction-following discipline and basic Ansible conventions.

Table 21 summarizes the key error types observed across models for this task.

### G.1.2 Policy configuration: 43065965

We analyzed the policy configuration task (Task ID: 43065965) across nine models to assess how well they handle structured API requests, variable access, and output formatting. Despite being relatively straightforward, the task revealed consistent model deficiencies. Most models failed due to incorrect or undefined variables, improper templating logic, and omission of key steps like printing the response. Notably, even high-performing models such as LLaMA 3.1 8B and WizardCoder hallucinated logic or over-templatized their script. Table 22 summarizes the key error types for each model.

### G.1.3 Templating: Task ID 43628823

Templating tasks revealed consistent and widespread failure patterns across nearly all models. The core of this task revolves around checking for file existence and formatting output appropriately—yet many models exhibited significant challenges with YAML validity, variable access, and templating logic. Common problems included the use of undefined or out-of-scope variables, incorrect output formatting, unsupported parameters in file-finding modules, and incomplete or malformed playbooks.

Table 21: Common error categories observed for Networking based issue observed across evaluated open-source LLMs. A checkmark (✓) indicates the error type was frequently observed for that model in the analyzed tasks. Model names abbreviated for space; see Table 12 for full names. DeepSeek-R1-Qwen model omitted due to consistent structural failures.

| Model | Variable Issues | Prohibited Tasks Inclusion | Syntax Errors | Instruction Compliance Failure |
|---|---|---|---|---|
| Qwen-Coder-7B | ✓ | ✗ | ✓ | ✓ |
| CodeGemma-7B | ✓ | ✓ | ✓ | ✗ |
| CodeLLaMa-7B | ✓ | ✓ | ✓ | ✓ |
| CodeLLaMa-13B | ✓ | ✓ | ✗ | ✓ |
| DeepSeek-R1-LLama-8B | ✓ | ✓ | ✓ | ✓ |
| LLaMA-3.1-8B | ✓ | ✓ | ✓ | ✓ |
| LLaMA-3.2-3B | ✓ | ✓ | ✗ | ✗ |
| Phi-3.5 | ✓ | ✓ | ✗ | ✗ |
| Qwen-7B-Instruct | ✓ | ✗ | ✓ | ✗ |
| StarCoder2-7B | ✗ | ✗ | ✓ | ✓ |
| WizardCoder-15B | ✓ | ✓ | ✗ | ✓ |

Table 22: Common error categories observed for Task ID 43065965 (Policy Configuration)

| Model | Variable Issues | Syntax Errors | Template Errors in Loop Control | Output Issues | Instruction Compliance Failure | Incorrect Host |
|---|---|---|---|---|---|---|
| Qwen-Coder | ✓ | ✓ | ✗ | ✓ | ✗ | ✗ |
| Phi-3.5 | ✓ | ✓ | ✗ | ✓ | ✗ | ✗ |
| WizardCoder | ✗ | ✗ | ✓ | ✗ | ✗ | ✗ |
| LLaMA 3.1 8B | ✓ | ✗ | ✗ | ✗ | ✓ | ✓ |
| LLaMA 3.2 3B | ✓ | ✓ | ✓ | ✓ | ✗ | ✗ |
| CodeGemma 7B | ✓ | ✓ | ✓ | ✓ | ✗ | ✓ |
| StarCoder 2 7B | ✗ | ✓ | ✓ | ✓ | ✓ | ✗ |
| CodeLLaMa 7B | ✓ | ✗ | ✗ | ✓ | ✓ | ✗ |
| CodeLLaMa 13B | ✓ | ✓ | ✓ | ✓ | ✓ | ✗ |

Models such as WizardCoder and Vicuna consistently failed due to incorrect output formats and improper attribute access. Instruction-tuned models like CodeLLaMa and LLaMA-3.1/3.2 also struggled with hallucinating unsupported attributes or creating invalid YAML. While models like CodeGemma and Phi-3.5 demonstrated better intent alignment, they often failed due to weak error handling, incorrect assumptions about Ansible variable structures, or incomplete implementations. Notably, DeepSeek-R1-Distil-Qwen-7B failed to generate even syntactically valid playbooks, suggesting a lack of training exposure to structured IaC generation. Table 23 summarizes the dominant error types observed across the models.

### G.1.4 Deployment Pipeline: Task ID 63688612

This task evaluates whether models can configure a multi-step deployment process by writing outputs to files on Linux-based compute nodes. A striking trend across nearly all models was the repeated misuse of Windows-specific Ansible modules (e.g., `win_stat`, `win_copy`) in a clearly defined Linux environment. This mistake appeared even in top-tier models like LLaMA-3.1-8B and CodeGemma-7B, indicating possible memorization from Windows-heavy training data.

Additionally, a number of models—such as Phi-3.5 and Qwen-Coder—failed due to logic errors where file write operations were skipped, or unsupported Ansible patterns like applying `line_in_file` on non-existent files. Some models hallucinated variables or services (Qwen-Coder), or misunderstood role delegation instructions. Others, such as DeepSeek-R1 variants and StarCoder2, produced incomplete or structurally incoherent playbooks, with no semantic alignment to the task description. WizardCoder notably failed to follow the code formatting template, blending text and code inappropriately. Overall, this task highlighted not just module-level errors, but also broader issues with infrastructure assumptions and procedural execution logic. Table 24 summarizes the key model-specific failure patterns.

### G.1.5 Server Configuration: Task Id- 44540690

Error trend actoss model can be found in Table 25.

Table 23: Common error categories observed for Task ID 43628823 (Templating)

| Model | Variable Issues | Attr/Param Issues | Output Issues | Playbook Incompleteness | Syntax Errors | Error Handling Failures |
|---|---|---|---|---|---|---|
| CodeGemma-7B | ✓ | ✓ | ✗ | ✗ | ✗ | ✓ |
| CodeLLaMa-7B | ✓ | ✓ | ✓ | ✗ | ✓ | ✓ |
| CodeLLaMa-13B | ✓ | ✓ | ✓ | ✓ | ✗ | ✗ |
| LLaMA-3.1-8B | ✗ | ✓ | ✓ | ✓ | ✓ | ✗ |
| Phi-3.5 | ✗ | ✓ | ✗ | ✗ | ✓ | ✓ |
| Qwen-2.5-7B | ✗ | ✗ | ✓ | ✓ | ✓ | ✓ |
| DeepSeek-R1-LLama-8B | ✗ | ✗ | ✗ | ✓ | ✓ | ✗ |
| DeepSeek-R1-Qwen-7B | ✗ | ✗ | ✗ | ✗ | ✗ | ✗ |
| LLaMA-3.2-3B | ✓ | ✓ | ✓ | ✗ | ✓ | ✗ |
| StarCoder2-7B | ✓ | ✓ | ✓ | ✓ | ✗ | ✗ |
| Vicuna-7B | ✓ | ✓ | ✓ | ✗ | ✓ | ✓ |
| WizardCoder-15B | ✗ | ✓ | ✓ | ✗ | ✗ | ✓ |

Table 24: Common error categories observed for Task ID 63688612 (Deployment Pipeline), for this task wizardcoder Do not follow code creation template so mix text and code.

| Model | Win Module Misuse | Invalid YAML / Syntax | Incorrect Logic / Skipped Tasks | Nonexistent File Write | Unsupported Params / Attr | Incomplete Playbook |
|---|---|---|---|---|---|---|
| CodeGemma-7B | ✓ | ✓ | ✗ | ✗ | ✓ | ✗ |
| CodeLLaMa-7B | ✓ | ✓ | ✓ | ✓ | ✓ | ✗ |
| DeepSeek-R1-LLama-8B | ✓ | ✓ | ✗ | ✗ | ✗ | ✓ |
| DeepSeek-R1-Qwen-7B | ✓ | ✓ | ✗ | ✗ | ✗ | ✓ |
| LLaMA-3.1-8B | ✓ | ✗ | ✓ | ✓ | ✗ | ✗ |
| LLaMA-3.2-3B | ✓ | ✗ | ✗ | ✗ | ✗ | ✗ |
| Phi-3.5 | ✗ | ✗ | ✓ | ✓ | ✗ | ✗ |
| Qwen-2.5-7B | ✗ | ✗ | ✗ | ✓ | ✗ | ✗ |
| Qwen-Coder-7B | ✓ | ✓ | ✓ | ✓ | ✓ | ✗ |
| StarCoder2-7B | ✓ | ✗ | ✗ | ✗ | ✗ | ✓ |
| WizardCoder-15B | ✗ | ✗ | ✗ | ✗ | ✗ | ✗ |

### G.1.6 File Management: Task Id

Error trend across models can be found in Table 26.

Table 25: Categorization of Common Errors in Playbook Generation Tasks

| Model | Invalid Yaml/Syntax | Wrong Module | Path/File Errors | Variable Issues | Attr/Param Errors | Delegation Errors |
|---|---|---|---|---|---|---|
| CodeGemma-7B | ✗ | ✓ | ✓ | ✓ | ✓ | ✓ |
| CodeLLaMa-7B | ✓ | ✗ | ✓ | ✗ | ✓ | ✓ |
| CodeLLaMa-13B | ✓ | ✗ | ✓ | ✓ | ✓ | ✓ |
| DeepSeek-R1-LLama-8B | ✓ | ✓ | ✗ | ✗ | ✗ | ✗ |
| DeepSeek-R1-Qwen-7B | ✓ | ✗ | ✗ | ✗ | ✗ | ✗ |
| DeepSeekCoder-v2 | ✓ | ✓ | ✓ | ✗ | ✓ | ✗ |
| LLaMA-3.1-8B | ✓ | ✓ | ✓ | ✗ | ✗ | ✗ |
| LLaMA-3.2-3B | ✓ | ✓ | ✓ | ✓ | ✗ | ✗ |
| Phi-3.5 | ✗ | ✓ | ✓ | ✓ | ✗ | ✗ |
| Qwen-7B-Instruct | ✓ | ✓ | ✓ | ✗ | ✓ | ✗ |
| Qwen-Coder-7B | ✓ | ✓ | ✓ | ✓ | ✓ | ✗ |
| StarCoder2-7B | ✗ | ✓ | ✓ | ✓ | ✓ | ✗ |
| WizardCoder-15B | ✓ | ✗ | ✗ | ✗ | ✗ | ✓ |

Table 26: Error Categorization in File Management Task

| Model | Invalid Yaml/Syntax | Wrong Host | Path/File Errors | Variable Issues | Attr/Param Errors | Templating Errors | Dir Creation Errors | Incomplete Playbook |
|---|---|---|---|---|---|---|---|---|
| CodeGemma | ✓ | ✗ | ✓ | ✓ | ✓ | ✓ | ✗ | ✗ |
| CodeLLaMa-7B | ✗ | ✗ | ✓ | ✗ | ✗ | ✗ | ✓ | ✓ |
| CodeLLaMa-13B | ✓ | ✓ | ✓ | ✗ | ✓ | ✗ | ✓ | ✓ |
| DeepSeek-R1-Distil-LLaMa-8B | ✓ | ✗ | ✗ | ✗ | ✗ | ✗ | ✗ | ✗ |
| DeepSeek-R1-Distil-Qwen-7B | ✓ | ✗ | ✗ | ✗ | ✗ | ✗ | ✗ | ✗ |
| DeepSeekCoder-v2 | ✗ | ✓ | ✓ | ✗ | ✗ | ✗ | ✗ | ✗ |
| LLaMa3.1-8B | ✓ | ✓ | ✓ | ✗ | ✗ | ✓ | ✓ | ✗ |
| LLaMa3.2-3B | ✓ | ✓ | ✗ | ✗ | ✗ | ✗ | ✗ | ✓ |
| Phi-3.5-Mini | ✓ | ✓ | ✓ | ✗ | ✓ | ✗ | ✗ | ✗ |
| Qwen-7B-Instruct | ✓ | ✗ | ✓ | ✗ | ✗ | ✓ | ✓ | ✗ |
| Qwen-Coder-7B-Instruct | ✓ | ✓ | ✓ | ✓ | ✓ | ✗ | ✗ | ✗ |
| Starcoder2 | ✗ | ✗ | ✓ | ✗ | ✓ | ✗ | ✗ | ✗ |
| WizardCoder-15B | ✗ | ✓ | ✓ | ✗ | ✗ | ✗ | ✗ | ✗ |